\documentclass[twoside]{article}
\usepackage{cl,epic,eepic,epsfig}
\title{Probabilistic top-down parsing and language modeling\\{\hspace*{1in}}\\
{\small Note that the page numbers and placement and formatting of figures\\
and tables differ from the version printed in the journal.\\\vspace*{-.1in}}}
\author{ Brian Roark%
         \thanks{Department of Cognitive and Linguistic Sciences, 
                 Box 1978, Brown University, 
                 Providence, RI  02912}\\
         \affil{Brown University}}
\runningtitle{\hspace*{1.5in}Top-down parsing}
\runningauthor{Roark}
\issue{27}{2}{2001}
\renewcommand{\Pr}{{\rm P}}
\newcommand{\Prhat}{{\widehat{\Pr}}}
\newcommand{\PrG}{\Pr_{\rm G}}
\newcommand{\scS}{{\mathcal{S}}}
\begin{document}
\maketitle
\begin{abstract}
This paper describes the functioning of a broad-coverage probabilistic
top-down parser, and its application to the problem of language
modeling for speech recognition.  The paper first introduces key
notions in language modeling and probabilistic parsing, and briefly
reviews some previous approaches to using syntactic structure for
language modeling.
A lexicalized probabilistic top-down parser is then presented, 
which performs very well, in terms of both the accuracy of returned
parses and the efficiency with which they are found, relative to the
best broad-coverage statistical parsers.  A new language model which utilizes 
probabilistic top-down parsing is then outlined, and empirical results
show that it improves upon previous work in test corpus perplexity.  
Interpolation with a trigram model yields an exceptional improvement
relative to the improvement observed by other models, demonstrating
the degree to which the information captured by our parsing model is
orthogonal to that captured by a trigram model.  A small recognition
experiment also demonstrates the utility of the model.
\end{abstract}
\section{Introduction}
With certain exceptions, computational linguists have in the
past generally formed a separate
research community from speech recognition researchers, despite some
obvious overlap of interest.  Perhaps one reason for this is that,
until relatively recently, few methods have come out of the natural 
language processing community that were shown to 
improve upon the very simple language models still standardly in use in 
speech recognition systems.  In the past few years, however, some
improvements have been made over these language models through
the use of statistical methods of natural language processing; and
the development of innovative, linguistically well-motivated
techniques for improving language models for speech recognition is
generating more interest among computational linguists.  While
language models built around shallow local dependencies are still the
standard in state-of-the-art speech recognition systems, there is
reason to hope that better language models can and will be developed
by computational linguists for this task.

This paper will examine language modeling for speech recognition from
a natural language processing point of view.  Some of the recent
literature investigating approaches that use syntactic structure in an  
attempt to capture long-distance dependencies for language modeling
will be reviewed.  A new language model, based on probabilistic
top-down parsing, will be outlined and compared with the previous
literature, and extensive empirical results will be presented which
demonstrate its utility.

Two features of our top-down parsing approach will emerge as key
to its success.  First, the top-down parsing algorithm builds a set of
{\it rooted} candidate parse trees from
left-to-right over the string, which allows it to calculate
a generative probability for each prefix string from the probabilistic
grammar, and hence a conditional probability for each word given the
previous words and the probabilistic grammar.  A left-to-right parser
whose derivations are not rooted, i.e. with derivations that can consist of
disconnected tree fragments, such as an LR or shift-reduce
parser, cannot incrementally calculate the probability of each prefix
string being generated by the probabilistic grammar, because their
derivations include probability mass from unrooted structures.  Only
at the point when their derivations become rooted (at the end of the
string) can generative string probabilities be calculated from the
grammar.  These parsers can calculate word probabilities based upon
the parser state -- as in \namecite{Chelba98a} -- but such a
distribution is not generative from the probabilistic grammar.

A parser that is not left-to-right, but which has rooted derivations,
e.g. a head-first parser, will be able to calculate generative joint
probabilities for entire strings; however it will not be able to
calculate probabilities for each word conditioned on previously
generated words, unless each derivation generates the words in the
string in exactly the same order.  For example, suppose that there are
two possible verbs that could be the head of a sentence.  For a
head-first parser, some derivations will have the first verb as the
head of the sentence, and the second verb will be generated after the
first; hence the second verb's probability will be conditioned on the
first verb. Other derivations will have the second verb as the head of
the sentence, and the first verb's probability will be conditioned on
the second verb.  In such a scenario, there is no way to decompose the
joint probability calculated from the set of derivations into the 
product of conditional probabilities using the chain rule.  Of course,
the joint probability can be used as a language model, but it cannot
be interpolated on a word-by-word basis with, say, a trigram model,
which we will demonstrate is a useful thing to do. 

Thus, our top-down parser allows for the incremental calculation of
generative conditional word probabilities, a property it shares with
other left-to-right parsers with rooted derivations such as Earley parsers
\cite{Earley70} or left-corner parsers \cite{Rosenkrantz70}.  

A second key feature of our approach is that top-down guidance
improves the efficiency of the search as more and more conditioning
events are extracted from the derivation for use in the probabilistic
model.  Because the rooted partial derivation is fully connected, all
of the conditioning information that might be extracted from the
top-down left context has already been specified, and a conditional
probability model built on this information will not impose any
additional burden on the search.  In contrast, an Earley or
left-corner parser will underspecify certain connections between
constituents in the left-context, and if some of the underspecified
information is used in the conditional probability model, it will have
to become specified.  Of course, this can be done, but at the expense
of search efficiency; the more that this is done, the less of a
benefit there is to be had from the underspecification.  A top-down
parser will, in contrast, derive an efficiency benefit from precisely
the information that is left underspecified in these other
approaches. 

Thus, our top-down parser makes it very easy to condition the
probabilistic grammar on an arbitrary number of values extracted from
the rooted, fully specified derivation.  This has lead us to a
formulation of the conditional probability model in terms of values
returned from tree-walking functions that themselves are contextually
sensitive.  The top-down guidance that is provided makes this approach
quite efficient in practice.

The following section will provide some background in probabilistic
context-free grammars and language modeling for speech recognition.
There will also be a brief review of previous work using syntactic
information for language modeling, before introducing our model in
section \ref{sec:tdp}.

\section{Background}
\subsection{Grammars and trees}\label{sec:gramm}
This section will introduce probabilistic (or stochastic) context-free
grammars (PCFGs)\footnote{For a detailed 
introduction to PCFGs, see e.g. \namecite{Manning99}.}, as well as
such notions as complete and partial parse trees, which will be
important in defining our language model later in the paper.  In
addition, we will explain some simple grammar transformations 
that will be used.  Finally, we will explain the notion of c-command,
which will be used extensively later as well.

PCFGs model the syntactic combinatorics of a language by extending
conventional context-free grammars (CFGs).  A CFG $G =
(V,T,P,S^\dag)$, consists of a set of non-terminal symbols
$V$, a set of terminal symbols $T$, a start symbol $S^\dag \in V$, and a
set of rule productions $P$ of the form: $A \rightarrow \alpha$, where
$\alpha \in (V \cup T)^{\ast}$. These context-free 
rules can be interpreted as saying that a non-terminal symbol $A$ 
expands into one\footnote{For ease of exposition, we will ignore
epsilon productions for now.  An epsilon production has the empty
string ($\epsilon$) on the
right-hand side, and can be written $A \rightarrow \epsilon$.
Everything that is said here can be straightforwardly extended to
include such productions.} or more either non-terminal or terminal
symbols, $\alpha$ = $X_{0} \ldots X_{k}$. A sequence of 
context-free rule expansions can be represented in a tree, with
parents expanding into one or more children below them in the tree.
Each of the individual local expansions in the tree is a rule
in the CFG.  Nodes in the tree with no children are
called leaves.  A tree whose leaves consist entirely of terminal
symbols is complete. Consider, for example, the parse tree shown in
figure \ref{fig:itree}(a):  the start symbol is $S^\dag$,
which expands into an S.  The S node expands
into an NP followed by a VP.  These non-terminal
nodes each in turn expand, and this process of expansion continues
until the tree generates the terminal string, ``\texttt{Spot chased the
ball}'', as leaves.

\begin{figure}[t]
\begin{picture}(106,145)(0,-125)
\put(25,15){(a)}
\put(34,-8){\small $S^\dag$}
\drawline(37,-12)(37,-22)
\put(34,-30){\small S}
\drawline(37,-34)(10,-44)
\put(3,-52){\small NP}
\drawline(10,-56)(10,-66)
\put(-0,-74){Spot}
\drawline(37,-34)(63,-44)
\put(56,-52){\small VP}
\drawline(63,-56)(41,-66)
\put(30,-74){\small VBD}
\drawline(41,-78)(41,-88)
\put(27,-96){chased}
\drawline(63,-56)(85,-66)
\put(78,-74){\small NP}
\drawline(85,-78)(73,-88)
\put(66,-96){\small DT}
\drawline(73,-100)(73,-110)
\put(66,-118){the}
\drawline(85,-78)(98,-88)
\put(90,-96){\small NN}
\drawline(98,-100)(98,-110)
\put(90,-118){ball}
\end{picture}
\begin{picture}(106,145)(-10,-125)
\put(51,15){(b)}
\put(60,-8){\small $S^\dag$}
\drawline(63,-12)(37,-22)
\put(34,-30){\small S}
\drawline(37,-34)(10,-44)
\put(3,-52){\small NP}
\drawline(10,-56)(10,-66)
\put(-0,-74){Spot}
\drawline(37,-34)(63,-44)
\put(56,-52){\small VP}
\drawline(63,-56)(41,-66)
\put(30,-74){\small VBD}
\drawline(41,-78)(41,-88)
\put(27,-96){chased}
\drawline(63,-56)(85,-66)
\put(78,-74){\small NP}
\drawline(85,-78)(73,-88)
\put(66,-96){\small DT}
\drawline(73,-100)(73,-110)
\put(66,-118){the}
\drawline(85,-78)(98,-88)
\put(90,-96){\small NN}
\drawline(98,-100)(98,-110)
\put(90,-118){ball}
\drawline(63,-12)(89,-22)
\put(80,-30){\small STOP}
\drawline(89,-34)(89,-44)
\put(80,-52){$\mathrm{\langle/s\rangle}$}
\end{picture}
\begin{picture}(119,145)(-40,-125)
\put(29,15){(c)}
\put(38,-8){\small $S^\dag$}
\drawline(41,-12)(22,-22)
\put(19,-30){\small S}
\drawline(22,-34)(10,-44)
\put(3,-52){\small NP}
\drawline(10,-56)(10,-66)
\put(-0,-74){Spot}
\drawline(22,-34)(34,-44)
\put(27,-52){\small VP}
\drawline(41,-12)(60,-22)
\put(51,-30){\small STOP}
\end{picture}
\caption{Three parse trees:
(a) a complete parse tree; (b) a complete 
parse tree with an explicit stop symbol; and (c) a partial
parse tree\label{fig:itree}}
\end{figure}

A CFG $G$ defines a language $L_G$, which is a subset of the set of
strings of terminal symbols, including only those that are leaves of
complete trees rooted at $S^\dag$, built with rules from the
grammar $G$.
We will denote strings either as $w$ or as $w_{0}w_{1}\ldots w_{n}$,
where $w_n$ is understood to be the last terminal symbol in the string.
For simplicity in displaying equations, from this point forward let
$w_{i}^{j}$ be the substring $w_{i}\ldots w_{j}$. 
Let $T_{w_0^n}$ be the set of all complete trees rooted at the start
symbol, with the string of terminals $w_{0}^{n}$ as
leaves.  We call $T_{w_0^n}$ the set of complete parses of $w_{0}^{n}$.  

A PCFG is a CFG with a probability assigned to each
rule;  specifically, each right-hand side has a probability given the
left-hand side of the rule.  The probability of a parse tree is
the product of the probabilities of each rule in the tree.  Provided a
PCFG is consistent (or tight), which it always will be in the
approach we will be advocating\footnote{A PCFG is consistent or tight
if there is no probability mass reserved for infinite trees.
\namecite{Chi98} proved that any PCFG estimated from a treebank with
the relative frequency estimator is tight.  All of the PCFGs that are
used in this paper are estimated using the relative frequency
estimator.}, this defines a proper probability distribution over completed
trees.  

A PCFG also defines a probability distribution over strings of
words (terminals) in the following way:
\begin{eqnarray}
\Pr(w_{0}^{n}) &=& \sum_{t \in T_{w_0^n}}\Pr(t) \label{eq:prst}
\end{eqnarray}
The intuition behind equation \ref{eq:prst} is that, if a string
is generated by the PCFG, then it will be produced
if and only if one of the trees in the set $T_{w_0^n}$ generated it.  Thus 
the probability of the string is the probability of the set $T_{w_0^n}$,
i.e. the sum of its members' probabilities.

Up to this point, we have been discussing strings of words without
specifying whether they are ``complete'' strings or not.  We will
adopt the convention that an explicit beginning of string symbol,
$\langle$s$\rangle$, and an explicit end symbol, $\langle$/s$\rangle$,
are part of the vocabulary, and a string $w_0^n$ is a complete string
if and only if $w_0$ is $\langle$s$\rangle$ and $w_n$ is
$\langle$/s$\rangle$.  Since the beginning of string symbol is not
predicted by language models, but rather is axiomatic in the 
same way that $S^\dag$ is for a parser, we can safely omit it
from the current discussion, and simply assume that it is there.  See
figure \ref{fig:itree}(b) for the explicit representation.

While a complete string of words must contain the end symbol as its
final word, a string prefix does not have this restriction.  For
example, ``\texttt{Spot chased the ball $\mathtt{\langle/s\rangle}$}'' is a
complete string, and the following is the set of prefix strings of
this complete string: ``\texttt{Spot}''; ``\texttt{Spot chased}'';
``\texttt{Spot chased the}''; ``\texttt{Spot chased the ball}''; and
``\texttt{Spot chased the ball $\mathtt{\langle/s\rangle}$}''.
A PCFG also defines a probability distribution over string
prefixes, and we will present this in terms of partial derivations.
A partial derivation (or parse) $d$ is defined with respect to a
prefix string $w_0^j$  as follows: it is the leftmost
derivation\footnote{A leftmost derivation is a derivation in which the
leftmost non-terminal is always expanded.} of
the string, with $w_j$ on the right-hand side of the last expansion in
the derivation.  Let $D_{w_0^j}$ be the set of all partial derivations
for a prefix string $w_0^j$.  Then 
\begin{eqnarray}
\Pr(w_0^j) &=& \sum_{d \in D_{w_0^j}}\Pr(d) \label{eq:pr_prst}
\end{eqnarray}

\begin{figure}[t]
\begin{picture}(188,210)(0,-190)
\put(91,15){(a)}
\put(97,-8){\small $S^\dag$}
\drawline(103,-12)(66,-22)
\put(63,-30){\small S}
\drawline(66,-34)(23,-44)
\put(16,-52){\small NP}
\drawline(23,-56)(23,-66)
\put(13,-74){Spot}
\drawline(66,-34)(109,-44)
\put(98,-52){\small S-NP}
\drawline(109,-56)(50,-66)
\put(43,-74){\small VP}
\drawline(50,-78)(14,-88)
\put(3,-96){\small VBD}
\drawline(14,-100)(14,-110)
\put(-0,-118){chased}
\drawline(50,-78)(86,-88)
\put(66,-96){\small VP-VBD}
\drawline(86,-100)(45,-110)
\put(38,-118){\small NP}
\drawline(45,-122)(22,-132)
\put(15,-140){\small DT}
\drawline(22,-144)(22,-154)
\put(16,-162){the}
\drawline(45,-122)(68,-132)
\put(52,-140){\small NP-DT}
\drawline(68,-144)(47,-154)
\put(39,-162){\small NN}
\drawline(47,-166)(47,-176)
\put(39,-184){ball}
\drawline(68,-144)(90,-154)
\put(65,-162){\small NP-DT,NN}
\drawline(90,-166)(90,-176)
\put(88,-184){$\epsilon$}
\drawline(86,-100)(127,-110)
\put(99,-118){\small VP-VBD,NP}
\drawline(127,-122)(127,-132)
\put(125,-140){$\epsilon$}
\drawline(109,-56)(168,-66)
\put(148,-74){\small S-NP,VP}
\drawline(168,-78)(168,-88)
\put(166,-96){$\epsilon$}
\drawline(103,-12)(140,-22)
\put(130,-30){\small $S^\dag$-S}
\drawline(140,-34)(140,-44)
\put(131,-52){$\langle$/s$\rangle$}
\end{picture}
\begin{picture}(288,210)(0,-190)
\put(91,15){(b)}
\put(97,-8){\small $S^\dag$}
\drawline(103,-12)(66,-22)
\put(63,-30){\small S}
\drawline(66,-34)(23,-44)
\put(16,-52){\small NP}
\drawline(23,-56)(23,-66)
\put(13,-74){Spot}
\drawline(66,-34)(109,-44)
\put(98,-52){\small S-NP}
\drawline(103,-12)(140,-22)
\put(130,-30){\small $S^\dag$-S}
\end{picture}
\caption{Two parse trees: (a) a complete left-factored parse tree with
epsilon productions and an explicit stop symbol; and (b) a partial
left-factored parse tree}\label{fig:ftree}
\end{figure}

We left-factor the PCFG, so that all productions are binary,
except those with a single terminal on the right-hand side and epsilon
productions\footnote{The only $\epsilon$-productions that we will use
in this paper are those introduced by left factorization.}. We do this
because it delays predictions about what non-terminals we expect
later in the string until we have seen more of the string.  In effect,
this is an underspecification of some of the predictions that our
top-down parser is making about the rest of the string.  The
left-factorization transform that we use is identical to what is
called right binarization in \namecite{Roark99b}.  See that paper for
more discussion of the benefits of factorization for top-down and
left-corner parsing.  For a grammar $G$, we define a factored grammar
$G_f$ as follows: 
\newcounter{ylist}
\begin{list}{\roman{ylist}.}{\setlength{\itemsep}{0in} \usecounter{ylist}}
\item ($A \rightarrow B$ \ $A$-$B$) $\in G_f$ iff ($A \rightarrow B\beta$) $\in G$, s.t. $B \in V$ and $\beta\in V^\ast$ 
\item ($A$-$\alpha \rightarrow B$ \ $A$-$\alpha B$) $\in G_f$
iff ($A \rightarrow \alpha B\beta$) $\in G$, s.t. $B \in V$, $\alpha\in V^+$, and $\beta\in V^\ast$ 
\item ($A$-$\alpha B \rightarrow \epsilon$) $\in G_f$ iff ($A$
$\rightarrow \alpha B$) $\in G$, s.t. $B \in V$ and $\alpha\in V^\ast$ 
\item ($A \rightarrow a$) $\in G_f$ iff ($A \rightarrow a$) $\in G$, s.t. $a \in T$
\end{list}
We can see the effect of this transform on our example parse trees in
figure \ref{fig:ftree}.  This underspecification of the non-terminal
predictions (e.g. VP-VBD in the example in figure \ref{fig:ftree}, as
opposed to NP), allows lexical items to become part of the
left-context, and so be used to condition production probabilities,
even the production probabilities of constituents that dominate them
in the unfactored tree.  It also brings words further downstream into
the look-ahead at the point of specification.  Note that partial trees
are defined in exactly the same way (figure \ref{fig:ftree}b), but
that the non-terminal yields are made up exclusively of the composite
non-terminals introduced by the grammar transform.  

This transform has a couple of very nice properties.  First, it is
easily reversible, i.e. every parse tree built with $G_f$ corresponds
to a unique parse tree built with $G$.  Second, if we use the relative
frequency estimator for our production probabilities, the probability
of a tree built with $G_f$ is identical to the probability of the
corresponding tree built with $G$.  

Finally, let us introduce the term \texttt{c-command}.  We will use
this notion in our conditional probability model, and it is also
useful for understanding some of the previous work in this area.  
The simple definition of c-command that we will be using in this paper
is the following:  a node $A$ c-commands a node $B$ if and 
only if (i) $A$ does not dominate\footnote{A node $A$ dominates a node $B$
in a tree if  and only if either (i) $A$ is the parent of $B$; or (ii) $A$
is the parent 
of a node $C$ that dominates $B$.} $B$; and (ii) the lowest branching node
(i.e. non-unary node) that dominates $A$ also dominates $B$.  Thus in
figure \ref{fig:itree}(a), the subject NP and the VP
each c-command the other, because neither dominates the other and the
lowest branching node above both (the S) dominates the
other.  Notice that the subject NP c-commands the object
NP, but not vice versa, since the lowest branching node that
dominates the object NP is the VP, which does not
dominate the subject NP.

\subsection{Language modeling for speech recognition}
This section will briefly introduce language modeling for statistical
speech recognition\footnote{For a detailed introduction to statistical
speech recognition, see \namecite{Jelinek97}.}.

In language modeling, we assign probabilities to strings of words.
To assign a probability, the chain rule is generally invoked.  
The chain rule states, for a string of $k$+1 words: 
\begin{eqnarray}
\Pr(w_{0}^{k}) &=&
\Pr(w_{0})\prod_{i=1}^{k}\Pr(w_{i}|w_{0}^{i-1})\label{eq:chain}
\end{eqnarray}
A Markov language model of order $n$ truncates the conditioning
information in the chain rule to include only the previous $n$ words.
\begin{eqnarray}
\Pr(w_{0}^{k}) &=&
\Pr(w_{0})\Pr(w_{1}|w_{0})\ldots \Pr(w_{n-1}|w_{0}^{n-2})\prod_{i=n}^{k}\Pr(w_{i}|w_{i-n}^{i-1})
\end{eqnarray}
These models are commonly called {\it n-gram\/} models\footnote{The
$n$ in {\it n-gram\/} is one more than the order of the Markov model,
since the $n$-gram includes the word being conditioned.}.
The standard language model used in many speech recognition systems is
the trigram model, i.e. a Markov model of order 2, which can be
characterized by the following equation:
\begin{eqnarray}
\Pr(w_{0}^{n-1}) &=&
\Pr(w_{0})\Pr(w_{1}|w_{0})\prod_{i=2}^{n-1}\Pr(w_{i}|w_{i-2}^{i-1})
\end{eqnarray}

To smooth the trigram models that are used in this paper, we
interpolate the probability estimates of higher order Markov models
with lower order Markov models \cite{Jelinek80}.  The idea behind
interpolation is simple and has been shown to be very effective.  For
an interpolated ($n$+1)-gram:
\begin{eqnarray}
\Pr(w_{i}|w_{i-n}^{i-1}) &=&
\lambda_{n}(w_{i-n}^{i-1})\Prhat(w_{i}|w_{i-n}^{i-1}) +
(1-\lambda_{n}(w_{i-n}^{i-1}))\Pr(w_{i}|w_{i-n+1}^{i-1}) \label{eq:int}
\end{eqnarray}
Here $\Prhat$ is the empirically observed relative frequency, and
$\lambda_{n}$ is a function from $V^{n}$ to [0,1].  This
interpolation is recursively applied to the smaller order $n$-grams
until the bigram is finally interpolated with the unigram,
i.e. $\lambda_{0}$ = 1.

\section{Previous work}
There have been attempts to jump over adjacent words to words farther
back in the left-context, without the use of dependency links or
syntactic structure, for example \namecite{Saul97} and Rosenfeld
\shortcite{Rosenfeld96,Rosenfeld97}.  We will focus our very brief review,
however, on those which use grammars or parsing for their language
models.  These can be divided into two rough groups:  those that use
the grammar as a language model; and those that use a parser to
uncover phrasal heads standing in an important relation (c-command) to
the current word.  The approach that we will subsequently present uses
the probabilistic grammar as 
its language model, but only includes probability mass from those
parses that are found, i.e. it uses the parser to find a subset of the
total set of parses (hopefully most of the high probability parses)
and uses the sum of their probabilities as an estimate of the true
probability given the grammar.

\subsection{Grammar models}
As mentioned in section \ref{sec:gramm}, a PCFG defines a
probability distribution over strings of words.  One approach to
syntactic language modeling is to use this distribution directly as a
language model.  There are efficient algorithms in the literature
\cite{Jelinek91,Stolcke95} for calculating exact string prefix
probabilities given a PCFG.  The algorithms both utilize a left-corner
matrix, which can be calculated in closed form through matrix
inversion.  They are limited, therefore, to grammars where the
non-terminal set is small enough to permit inversion.  String prefix
probabilities can be straightforwardly used to compute conditional
word probabilities by definition: 
\begin{eqnarray}
\Pr(w_{j+1}|w_0^j) &=& \frac{\Pr(w_0^{j+1})}{\Pr(w_0^j)}\label{eq:wd_prst}
\end{eqnarray}

\namecite{Stolcke94} and \namecite{Jurafsky95} used these basic ideas
to estimate bigram probabilities from hand-written PCFGs,
which were then used in language models.  Interpolating the observed
bigram probabilities with these calculated bigrams led, in both cases, 
to improvements in word error rate over using the observed bigrams
alone, demonstrating that there is some benefit to using these syntactic
language models to generalize beyond observed $n$-grams.  

\subsection{Finding phrasal heads}\label{sec:SLM}
Another approach that uses syntactic structure for language modeling
has been to use a shift-reduce parser to ``surface'' 
c-commanding phrasal head words or part-of-speech (POS) tags from 
arbitrarily far back in the prefix string, for use in a trigram-like
model.  

A shift-reduce parser\footnote{For details, see
e.g. \namecite{Hopcroft79}.} operates from left-to-right using a stack and
a pointer to the next word in the input string.  Each stack entry
consists minimally of a non-terminal label.  The parser performs two
basic operations: (i) {\it shifting\/}, which involves pushing the
POS label of the next word onto the stack
and moving the 
pointer to the following word in the input string; and (ii) {\it
reducing\/}, which takes the top $k$ stack entries and replaces them
with a single new entry, the non-terminal label of which is the left-hand
side of a rule in the grammar which has the $k$ top stack entry labels on
the right-hand side.  For example, if there is a rule NP
$\rightarrow$ DT NN, and the top two stack entries are NN and DT,  
then those two entries can be popped off of the stack and an entry
with the label NP pushed onto the stack.  

\namecite{Goddeau92} used a robust deterministic shift-reduce parser
to condition word probabilities by extracting a specified number of
stack entries from the top of the current state, and conditioning on
those entries in a way similar to an $n$-gram.  In empirical trials,
Goddeau used the top 2 stack entries to condition the word
probability.  He was able to reduce both sentence and word error rates
on the ATIS corpus using this method.

The ``Structured Language Model'' (SLM) used in Chelba and
Jelinek \shortcite{Chelba98a,Chelba98b,Chelba99},
\namecite{Jelinek99}, and \namecite{Chelba00} is similar to that of
Goddeau, except that (i) their shift-reduce parser follows a
non-deterministic beam 
search, and (ii) each stack entry contains, in addition to the
non-terminal node label, the head-word of the constituent.  The
SLM is like a trigram, except that the conditioning words
are taken from the tops of the stacks of candidate parses in the beam,
rather than from the linear order of the string.

Their parser functions in three stages.  The first stage assigns a
probability to the word given the left-context (represented by the
stack state).  The second stage
predicts the POS given the word and the left-context.  The
last stage performs all possible parser operations (reducing stack
entries and shifting the new word).  When there is no more parser work 
to be done (or, in their case, when the beam is full), the following
word is predicted.  And so on until the end of the string.

Each different POS assignment or parser operation is
a step in a derivation.  Each distinct derivation path within the beam 
has a probability and a stack state associated with it.  Every stack
entry has a non-terminal node label and a designated head
word of the constituent.  When all of the
parser operations have finished at a particular point in the string,
the next word is predicted as follows.  For each derivation in the
beam, the head words of the two
topmost stack entries form a trigram with the conditioned word.  This
interpolated trigram probability is then multiplied by the normalized
probability of the derivation, to provide that derivation's
contribution to the probability of the word.  More precisely, for a
beam of derivations $D_{i}$
\begin{eqnarray}
\Pr(w_{i+1}|w_0^i) &=& \frac{\sum_{d \in D_{i}}
\Pr(w_{i+1}|h_{0d},h_{1d}) \Pr(d)}{\sum_{d \in D_{i}}\Pr(d)}
\end{eqnarray}
where $h_{0d}$ and $h_{1d}$ are the lexical heads of the top two
entries on the stack of $d$.

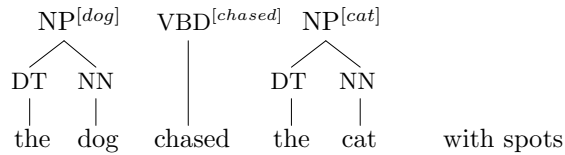
\begin{figure}[t]
\begin{picture}(106,70)(-80,-110)
\put(5,-52){NP$^{[dog]}$}
\drawline(15,-56)(2,-66)
\put(-5,-74){\small DT}
\drawline(2,-78)(2,-88)
\put(-4,-96){the}
\drawline(15,-56)(27,-66)
\put(20,-74){\small NN}
\drawline(27,-78)(27,-88)
\put(20,-96){dog}
\put(50,-52){\small VBD$^{[chased]}$}
\drawline(62,-56)(62,-88)
\put(49,-96){chased}
\put(105,-52){NP$^{[cat]}$}
\drawline(114,-56)(101,-66)
\put(93,-74){\small DT}
\drawline(101,-78)(101,-88)
\put(94,-96){the}
\drawline(114,-56)(127,-66)
\put(119,-74){\small NN}
\drawline(127,-78)(127,-88)
\put(120,-96){cat \hspace*{.3in}with spots}
\end{picture}
\caption{Tree representation of a derivation state}\label{fig:derst}
\end{figure}

Figure \ref{fig:derst} gives a partial tree representation of a
potential derivation state for the string ``\texttt{the dog chased the
cat with spots}'', at the point when the word ``\texttt{with}'' is to be
predicted.  The shift-reduce parser will have, perhaps, built the
structure shown, and the stack state will have an NP entry with the
head ``\texttt{cat}'' at the top of the stack, and a VBD entry with
the head ``\texttt{chased}'' second on the stack.  In the Chelba and
Jelinek model, the probability of ``\texttt{with}'' is conditioned on
these two head words, for this derivation.

Since the specific results of the SLM will be compared in detail with
our model when the empirical results are presented, at this point we
will simply state that they have achieved a reduction in both perplexity
and WER over a standard trigram using this model.

The rest of this paper will present our parsing model, its application
to language modeling for speech recognition, and empirical results.
\pagebreak
\section{Top-down parsing and language modeling}\label{sec:tdp}
Statistically-based heuristic best-first or beam-search strategies
\cite{Caraballo98,Charniak98,Goodman97} have yielded an
enormous improvement in the quality and speed of parsers, even without
any guarantee that the parse returned is, in fact, that with the
maximum likelihood for the probability model.  The parsers with the
highest published broad-coverage parsing accuracy, which include
Charniak \shortcite{Charniak97,Charniak00}, Collins
\shortcite{Collins97,Collins99}, and \namecite{Ratna97} all 
utilize simple and straightforward statistically-based search
heuristics, pruning the search space quite
dramatically\footnote{\namecite{Johnson99}, \namecite{Henderson99},
and \namecite{Collins00} demonstrate methods for choosing the best
complete parse tree from among a set of complete parse trees, and the
latter two show accuracy improvements over some of the parsers cited
above, from which they generated their candidate sets.  Here I will be
comparing my work with parsing algorithms, i.e. algorithms which build
parses for strings of words.}.  
Such methods are nearly always used in conjunction with some form of
dynamic programming (henceforth DP).  That is, search
efficiency for these parsers is improved 
by both statistical search heuristics and DP.  Here we will
present a parser that uses simple search heuristics of this sort 
without DP.  Our approach is found to yield very
accurate parses efficiently, and, in addition, to lend itself
straightforwardly to estimating word probabilities on-line, i.e. in
a single pass from left-to-right.  This on-line characteristic allows
our language model to be interpolated on a word-by-word basis with
other models, such as the trigram, yielding further improvements.

Next we will outline our conditional probability model over rules in
the PCFG, followed by a presentation of the top-down parsing
algorithm. We
will then present empirical results in two domains: one to compare with
previous work in the parsing literature, and the other to compare with
previous work using parsing for language modeling for speech
recognition, in particular with the Chelba and Jelinek results
mentioned above.

\subsection{Conditional probability model}
A simple PCFG conditions rule probabilities on the
left-hand side of the rule.  It has been shown repeatedly --
e.g. \namecite{Briscoe93}, \namecite{Charniak97},
\namecite{Collins97}, \namecite{Inui97}, \namecite{Johnson98b} --
that conditioning the probabilities of structures on the
context within which they appear, for example on the lexical head of a
constituent \cite{Charniak97,Collins97}, on the label of its parent
non-terminal \cite{Johnson98b}, or, ideally, on both and many other
things besides, leads to a much better parsing model and results in
higher parsing accuracies.

One way of thinking about conditioning the probabilities of
productions on contextual information, e.g. the label of the parent of
a constituent or the lexical heads of constituents, is as annotating
the extra conditioning information onto the labels in 
the context-free rules.  Examples of this are bilexical grammars -- see
e.g. \namecite{Eisner99}, \namecite{Charniak97}, \namecite{Collins97} --
where the lexical heads of each constituent are annotated on both the
right- and left-hand sides of the context free rules, under the
constraint that every constituent inherits the lexical head from
exactly one of its children, and the lexical head of a POS is its
terminal item.  Thus the rule S $\rightarrow$ NP VP becomes, for
instance, S[{\it barks}] $\rightarrow$ NP[{\it dog}] VP[{\it barks}].
One way to estimate the probabilities of these rules is to annotate
the heads onto the constituent labels
in the training corpus, and simply count the number of times
particular productions occur (relative frequency estimation). This
procedure yields conditional probability distributions of 
constituents on the right-hand side with their lexical heads, given
the left-hand side constituent and 
its lexical head.  The same procedure works if we annotate parent
information onto constituents.  This is how \namecite{Johnson98b}
conditioned the probabilities of productions: the left-hand side is no
longer, for example, S, but rather S$^\uparrow$SBAR,
i.e. an S with SBAR as parent.  Notice, however, that in
this case the annotations on the right-hand side are predictable from
the annotation on the left-hand side (unlike, for example, bilexical
grammars), so that the relative frequency estimator yields conditional
probability distributions of the original rules, given the parent of
the left-hand side.  

All of the conditioning information that we will be considering will
be of this latter sort: the only novel predictions being made by rule
expansions are the 
node-labels of the constituents on the right-hand side.  Everything
else is already specified by the left-context.  We use the relative
frequency estimator, and smooth our production probabilities by
interpolating the relative frequency estimates with those obtained by
``annotating'' less contextual information.

This perspective on conditioning production probabilities makes it
easy to see that, in essence, by conditioning these probabilities, we
are growing the state space.  That is, the number of distinct
non-terminals grows to include the composite labels; so does the
number of distinct productions in the grammar.  In a top-down parser,
each rule expansion 
is made for a particular candidate parse, which carries with it the
entire rooted derivation to that point; in a sense, the
left-hand side of the rule is annotated with the entire left-context,
and the rule probabilities can be conditioned on any aspect of this
derivation.

We do not use the entire left-context to condition the rule
probabilities, but rather ``pick-and-choose'' which events in the 
left-context we would like to condition on.  One can think of the
conditioning events as functions, which take the partial tree
structure as an argument and return a value, upon which the rule
probability can be conditioned.  Each of these functions is an
algorithm for walking the provided tree and returning a value.  
For example, suppose that we want to condition the probability of the
rule $A \rightarrow \alpha$.  We might write a function that takes the
partial tree, finds the parent of the left-hand side
of the rule and returns its node label.  If the left-hand side has no
parent, i.e. it is at the root of the tree, the function returns the
null value (NULL).  We might write another function that returns the
non-terminal label of the closest sibling to the left of $A$, and NULL
if no such node exists.  We can then condition the probability of the
production on the values that were returned by the set of functions.

Recall that we are working with a factored grammar, so some of the
nodes in the factored tree have non-terminal labels that were created by the
factorization, and may not be precisely what we want for conditioning
purposes.  In order to avoid
any confusions in identifying the non-terminal label of a particular
rule production in either its factored or non-factored version, we
introduce the function \texttt{constituent(}$A$\texttt{)\/} for every
non-terminal in the factored grammar $G_f$, which is simply the label of the 
constituent whose factorization results in $A$.  For example, in figure
\ref{fig:ftree}, \texttt{constituent(}NP-DT-NN\texttt{)\/} is simply 
NP.

Note that a function can return different values depending upon the
location in the tree of the non-terminal that is being expanded.  For
example, suppose that we have a function that returns the label of the
closest sibling to the left of \texttt{constituent(}$A$\texttt{)\/} or
NULL if no such node exists.  Then a subsequent function could be
defined as follows:  return the parent of the parent (the grandparent)
of \texttt{constituent(}$A$\texttt{)\/} {\it only if\/} 
\texttt{constituent(}$A$\texttt{)\/} has no sibling to the left -- in
other words, if the previous function returns NULL;
{\it otherwise\/} return the 2nd closest sibling to the left of
\texttt{constituent(}$A$\texttt{)\/}, or, as always, NULL if no such
node exists.  If the function returns, for example, ``NP'', this could
either mean that the grandparent is NP or the 2nd closest sibling is
NP; yet there is no ambiguity in the meaning of the function,
since the result of the previous function disambiguates between the
two possibilities.  

\begin{figure}[t]
\begin{center}
\begin{picture}(306,236)(0,-236)
\put(-30,-10){\underline{\bf For all rules $A \rightarrow \alpha$}}
\begin{footnotesize}
\put(144,-7){\circle{6}}
\put(142.5,-8.5){\tiny 0}
\put(150,-10){$A$}
\drawline(153,-12)(153,-23)
\put(69,-27){\circle{6}}
\put(67.5,-28.5){\tiny 1}
\put(75,-30){the parent, $Y_p$, of \texttt{constituent(}$A$\texttt{)\/} in the derivation}
\drawline(153,-32)(153,-43)
\put(49,-47){\circle{6}}
\put(47.5,-48.5){\tiny 2}
\put(55,-50){the closest sibling, $Y_s$, to the left of \texttt{constituent(}$A$\texttt{)\/} in the derivation}
\drawline(153,-52)(65,-83)
\drawline(153,-52)(235,-83)
\put(205,-70){\scriptsize\it $A$ = POS, $Y_s \neq$ NULL}
\put(-6,-87){\circle{6}}
\put(-7.5,-88.5){\tiny 3}
\put(0,-90){the parent, $Y_g$, of $Y_p$ in the
derivation}
\put(225,-90){the closest c-commanding}
\put(235,-97){lexical head to $A$}
\drawline(275,-100)(275,-111)
\put(220,-120){the next closest c-commanding}
\put(235,-127){lexical head to $A$}

\drawline(55,-93)(25,-121)
\drawline(55,-93)(135,-121)
\put(105,-110){\scriptsize\it $A$ = POS}
\put(-11,-137){\circle{6}}
\put(-12.5,-138.4){\tiny 4}
\put(0,-130){the closest sibling,}
\put(-5,-141){$Y_{ps}$, to the left of
$Y_p$}
\put(105,-130){the POS of the closest}
\put(95,-141){c-commanding lexical head to $A$}
\drawline(35,-143)(35,-160)
\put(-16,-172){\circle{6}}
\put(-17.5,-173.5){\tiny 5}
\put(-10,-165){If $Y_s$ is CC, the leftmost child}
\put(-10,-175){of the conjoining category; else NULL}
\drawline(165,-143)(225,-165)
\put(165,-176){the closest c-commanding lexical head to $A$}
\drawline(35,-180)(35,-195)
\put(-16,-207){\circle{6}}
\put(-17.5,-208.5){\tiny 6}
\put(-10,-200){the lexical head of \texttt{constituent(}$A$\texttt{)\/} if already seen;}
\put(-10,-210){otherwise the lexical head of the closest}
\put(-10,-220){constituent to the left of $A$ within \texttt{constituent(}$A$\texttt{)\/}}
\drawline(255,-185)(255,-200)
\put(225,-210){the next closest c-commanding}
\put(235,-220){lexical head to $A$}
\end{footnotesize}
\end{picture}
\end{center}
\caption{Conditional probability model represented as a decision tree,
identifying the location in the partial parse tree of the conditioning
information}\label{fig:code}
\end{figure}

The functions that were used for the present study to condition the
probability of the rule, $A \rightarrow \alpha$, are
presented in figure \ref{fig:code}, in a tree structure.  
This is a sort of decision tree for a tree-walking algorithm to decide
what value 
to return, for a given partial tree and a given depth.  For example,
if the algorithm is asked for the value at level 0, it will return $A$,
the left-hand side of the rule being expanded\footnote{Recall that $A$
can be a composite non-terminal introduced by grammar factorization.
When the function is defined in terms of
\texttt{constituent(}$A$\texttt{)\/}, the values returned are obtained
by moving through the non-factored tree.}.  Suppose the algorithm is
asked for the value at level 4.  After level 2 there is a branch in
the decision tree.  If the left-hand side of the rule is a POS, and 
there is no sibling to the left of \texttt{constituent(}$A$\texttt{)\/}
in the derivation, then the algorithm takes the right branch of the
decision tree to decide what value to return; otherwise
the left branch.  Suppose it takes the left branch.  Then after level
3, there is another branch in the decision tree.  If the left-hand
side of the production is a POS, then the algorithm takes the right
branch of the decision tree, and returns (at level 4) the POS of the
closest c-commanding lexical head to $A$, which it finds by walking the
parse tree; if the left-hand side of the rule is not a POS, then the
algorithm returns (at level 4) the closest sibling to the left of
the parent of \texttt{constituent(}$A$\texttt{)\/}.

The functions that we have chosen for this paper follow from the
intuition (and experience) that what helps parsing is different
depending on the constituent that is being expanded.  POS
nodes have lexical items on the right-hand side, and hence can bring
some of the head-head dependencies into the model that have been shown
to be so effective.  If the POS is leftmost within its constituent,
then very often the lexical item is sensitive to the governing
category to which it is attaching.  For example, if the POS is a
preposition, then its probability of expanding to a particular word is
very different if it is attaching to a noun phrase versus a verb
phrase, and perhaps quite different depending on the head of the
constituent to which it is attaching.  Subsequent POSs within
a constituent are likely to be open class words, and less
dependent on these sorts of attachment preferences.

Conditioning on parents and siblings of the left-hand side has proven
to be very useful.  To understand why this is the case, one need
merely to think of VP expansions.  If the parent of a 
VP is another VP (i.e. if an auxiliary or modal verb is
used), then the distribution over productions is different than if the
parent is an S.  Conditioning on head information, both
POS of the head and the lexical item itself, has proven
useful as well, although given our parser's left-to-right orientation,
in many cases the head has not been encountered within the particular
constituent.  In such a case, the head of the last child within the
constituent is 
used as a proxy for the constituent head.  All of our conditioning
functions, with one exception, return either parent or sibling node
labels at some specific distance from the left-hand side, or head
information from c-commanding constituents.  The exception is the
function at level 5 along the left branch of the tree in figure
\ref{fig:code}.  Suppose that 
the node being expanded is being conjoined with another node, which we
can tell by the presence or absence of a CC node.  In that
case, we want to condition the expansion on how the conjoining
constituent expanded.  In other words, this attempts to capture 
a certain amount of parallelism between the expansions of conjoined
categories. 

In presenting the parsing results, we will systematically vary the
amount of conditioning information, so as to get an idea of the
behavior of the parser.  We will refer to the amount of conditioning
by specifying the deepest level from which a value is returned for each
branching path in the decision tree, from left to right in figure
\ref{fig:code}: the first number is for left-contexts where the left
branch of the decision tree is always followed (non-
POS non-terminals on the left-hand side); the second number for a left
branch followed by a right branch (POS nodes that are
leftmost within their constituent); and the third number for the
contexts where the right branch is always followed (POS nodes
that are not leftmost within their constituent).  For
example, (4,3,2) would represent a conditional probability model that
(i) returns NULL for all functions below level four in all contexts;
(ii) returns NULL for all functions below level three if the
left-hand side is a POS; and (iii) returns NULL for all
functions below level two for non-leftmost POS expansions.

\begin{small}
\begin{center}
\begin{table*}[t]
\begin{tabular} {|p{.8in}|p{.8in}|p{3.2in}|}
\hline
{Conditioning} & {Mnemonic label} & {Information level}\\\hline 
{0,0,0} & {none} & {Simple PCFG}\\\hline
{2,2,2} & {par+sib} & {Small amount of structural context}\\\hline
{5,2,2} & {NT struct} & {All structural (non-lexical) context for non-POS}\\\hline
{6,2,2} & {NT head} & {Everything for non-POS expansions}\\\hline
{6,3,2} & {POS struct} & {More structural info for leftmost POS
expansions}\\\hline 
{6,5,2} & {attach} & {All attachment info for leftmost POS
expansions}\\\hline 
{6,6,4} & {all} & {Everything}\\\hline
\end{tabular}
\caption{Levels of conditioning information, mnemonic labels, and a
brief description of the information level for empirical results}\label{tab:meaning}
\end{table*}
\end{center}
\end{small}

Table \ref{tab:meaning} gives a breakdown of the different levels of
conditioning information used in the empirical trials, with a mnemonic
label that will be used when presenting results.  These different
levels were chosen as somewhat natural points at which to observe how
much of an effect increasing the conditioning information has.  We
first include structural information from the context, i.e. node
labels from constituents in the left context.  Then we add lexical
information, first for non-POS expansions, then for leftmost
POS expansions, then for all expansions.

All of the conditional probabilities are linearly interpolated.  For
example, the probability of a rule conditioned on six 
events is the linear interpolation of two probabilities: (i) the
empirically observed relative frequency of the rule when the six
events co-occur; and (ii) the probability of the rule conditioned on
the first five events (which is in turn interpolated).  The
interpolation coefficients are a function of the frequency of the set
of conditioning events, and are estimated by iteratively adjusting the
coefficients so as to maximize the likelihood of a held out corpus. 

This was an outline of the conditional probability model that we used
for the PCFG.  The model allows us to assign probabilities to
derivations, which can be used by the parsing algorithm to decide
heuristically which candidates are promising and should be expanded,
and which are less promising and should be pruned.
We now outline the top-down parsing algorithm.

\subsection{Top-down Probabilistic Parsing}
This parser is essentially a stochastic version of the top-down parser 
described in \namecite{Aho86}.  It uses a PCFG with a conditional
probability model of the sort defined in the previous section.  We
will first define {\it candidate analysis\/} (i.e. a partial parse),
and then a {\it derives\/} relation between candidate analyses.  We
will then present the algorithm in terms of this relation.

The parser takes an input string $w_0^n$, a PCFG $G$, and a priority queue of
candidate analyses.  A candidate analysis $C = (D,\scS,P_D,F,w_i^n)$
consists of a derivation $D$, a stack $\scS$, a derivation probability
$P_D$, a 
figure-of-merit $F$, and a string $w_i^n$ remaining to be parsed.  The
first word in the string remaining to be parsed, $w_i$, we will call the
{\it look-ahead\/} word. 
The derivation $D$ consists of a sequence of rules used from $G$.
The stack $\scS$ contains a sequence of non-terminal symbols, and an
end-of-stack marker \$ at the bottom.  The probability
$P_D$ is the product of 
the probabilities of all rules in the derivation $D$.
$F$ is the product of $P_D$ and a look-ahead probability,
LAP($\scS$,$w_i$), which is a measure of the likelihood
of the stack $\scS$ rewriting with $w_i$ at its left corner.

We can define a {\it derives\/} relation, denoted $\Rightarrow$,
between two candidate analyses as follows. $(D,\scS,P_D,F,w_i^n)
\Rightarrow
(D^\prime,\scS^\prime,P_{D^\prime},F^\prime,w_j^n)$ if
and only if\footnote{Again, for ease of exposition, we will ignore
$\epsilon$-productions.  Everything presented here can be
straightforwardly extended to include them. The + in (i) denotes
concatenation. To avoid confusion between sets and sequences,
$\emptyset$ will not be used for empty strings or sequences, rather
the symbol $\langle\rangle$ will be used. Note that the script $\scS$
is used to denote stacks, while $S^{\dag}$ is the start symbol.} 
\newcounter{yylist}
\begin{list}{\roman{yylist}.}{\setlength{\itemsep}{0in} \usecounter{yylist}}
\item $D^\prime = D\ +\ A \rightarrow X_0 \ldots X_k$
\item $\scS$ = $A\alpha$\$;
\item either $\scS^\prime = X_0 \ldots X_k\alpha$\$ and $j$ = $i$\\
or $k$ = 0, $X_0 = w_i$, $j$ = $i$+1, and $\scS^\prime = \alpha$\$;
\item $P_{D^\prime} = P_D\Pr(A \rightarrow
X_0 \ldots X_k)$; and
\item $F^\prime = P_{D^\prime}\mathrm{LAP}(\scS^\prime,w_j)$
\end{list}

The parse begins with a single candidate analysis on the priority queue:
($\langle\rangle$,$S^\dag$\$,1,1,$w_0^n$).  It then proceeds as
follows.  The top ranked candidate 
analysis, $C = (D,\scS,P_D,F,w_i^n)$, is popped from the priority queue.
If $\scS$ = \$ and $w_i$ = $\langle$/s$\rangle$, then the analysis is complete.
Otherwise, all $C^\prime$ such that $C \Rightarrow C^\prime$ are
pushed onto the priority queue.

We implement this as a beam search.  For each word position $i$, we have 
a separate priority queue, $H_i$, of analyses with look-ahead $w_i$. When there are
``enough'' analyses by some criteria (which we will discuss below) on
priority queue $H_{i+1}$, all candidate analyses remaining on $H_{i}$ are
discarded.  Since $w_n$ = $\langle$/s$\rangle$, all parses that are pushed onto
$H_{n+1}$ are complete.  The parse on $H_{n+1}$ with the
highest probability is returned for evaluation.  In the case that no
complete parse is found, a partial parse is returned and evaluated.  

The LAP is the probability of a particular terminal being the 
next left-corner of a particular analysis.  The terminal may be the
left-corner of the top-most non-terminal on the stack of the analysis
or it might be 
the left-corner of the {\it nth} non-terminal, after the top
$n$--1 non-terminals have rewritten to $\epsilon$. Of course, we
cannot expect to have adequate statistics for each non-terminal/word
pair that we encounter, so we smooth to the POS.  Since we do
not know the POS 
for the word, we must sum the LAP for all POS
labels\footnote{Equivalently, we can split the analyses at this point,
so that there is one POS per analysis.}.  

For a PCFG $G$, a stack $\scS = A_{0} \dots A_{n}$\$ (which we will write
$A_0^n$\$) and a look-ahead terminal item $w_i$, we define the look-ahead
probability as follows: 
\begin{equation}
\mathrm{LAP}(\scS,w_i) = \sum_{\alpha \in (V \cup T)^{*}} \PrG(A_0^n
\stackrel{\star}{\rightarrow} w_i\alpha)
\end{equation}  
We recursively estimate this with two empirically observed conditional
probabilities for every non-terminal $A_{i}$:
$\Prhat (A_{i} \stackrel{\star}{\rightarrow} w_i\alpha)$
and $\Prhat (A_{i} \stackrel{\star}{\rightarrow} \epsilon)$.
The same empirical probability, $\Prhat (A_{i} \stackrel{\star}{\rightarrow} X\alpha)$, is collected for every
pre-terminal $X$ as well. The LAP approximation for a given
stack state and look-ahead terminal is:
\begin{eqnarray}
\PrG(A_j^n \stackrel{\star}{\rightarrow} w_i\alpha) 
& \approx &  \PrG(A_{j}\stackrel{\star}{\rightarrow} w_i\alpha) + \Prhat (A_{j} \stackrel{\star}{\rightarrow} \epsilon)
\PrG(A_{j+1}^n \stackrel{\star}{\rightarrow}
w_i\alpha)  
\end{eqnarray}
where
\begin{equation}
\PrG(A_{j} \stackrel{\star}{\rightarrow} w_i\alpha) 
\approx  \lambda_{A_{j}} \Prhat (A_{j}
\stackrel{\star}{\rightarrow} w_i\alpha) + (1-\lambda_{A_{j}}) \sum_{X \in V}  \Prhat (A_{j}
\stackrel{\star}{\rightarrow} X\alpha) \Prhat (X
\rightarrow w_i) 
\end{equation}
The lambdas are a function of the frequency of the non-terminal
$A_{j}$, in the standard way \cite{Jelinek80}.

The beam threshold at word $w_{i}$ is a function of the
probability of the top ranked candidate analysis on priority queue $H_{i+1}$ and 
the number of candidates on $H_{i+1}$.  The basic idea is that we want 
the beam to be very wide if there are few analyses that have been
advanced, but relatively narrow if many analyses have been advanced.  If
$\tilde{p}$ is the probability of the highest ranked analysis on
$H_{i+1}$, then another analysis is discarded if its probability falls 
below $\tilde{p}f(\gamma,|H_{i+1}|)$, where $\gamma$ is an initial
parameter, which we call the {\it base beam factor\/}.  For the
current study,  $\gamma$ was $10^{-11}$, unless 
otherwise noted, and $f(\gamma,|H_{i+1}|) = \gamma|H_{i+1}|^{3}$.  Thus, 
if 100 analyses have already been pushed onto $H_{i+1}$, then a
candidate analysis must have a probability above $10^{-5}\tilde{p}$ 
to avoid being pruned.  After 1000
candidates, the beam has narrowed to $10^{-2}\tilde{p}$.  There is also a
maximum number of allowed analyses on $H_{i}$, in case the parse
fails to advance an analysis to $H_{i+1}$.  This was typically 10,000.

As mentioned in section \ref{sec:gramm}, we left-factor the grammar,
so that all productions are binary, except  
those with a single terminal on the right-hand side and epsilon
productions.  The only $\epsilon$-productions are those introduced by left
factorization.  Our factored grammar was produced
by factoring the trees in the training corpus before grammar
induction, which proceeded in the standard way, by counting rule
frequencies.
\pagebreak
\section{Empirical results}
The empirical results will be presented in three stages:  (i) trials to
examine the accuracy and efficiency of the parser; (ii) trials to examine
its effect on test corpus perplexity and recognition performance; and
(iii) trials to examine the effect of beam variation on these
performance measures.  Before presenting the results, we will
introduce the methods of evaluation.

\subsection{Evaluation}\label{sec:eval}
Perplexity is a standard measure within the speech recognition
community for comparing language models.  In principle, if two models
are tested on the same
test corpus, the model that assigns the lower perplexity to the 
test corpus is the model closest to the true distribution of the
language, and thus better as a prior model for
speech recognition.  Perplexity is the exponential of the cross
entropy, which we will define next.

Given a random variable $X$ with distribution $p$ and a probability
model $q$, the cross entropy, $H(p,q)$ is defined as follows: 
\begin{eqnarray}
H(p,q) &=& -\sum_{x \in X} p(x) \log q(x)
\end{eqnarray}
Let $p$ be the true distribution of
the language.  Then, under certain assumptions\footnote{See
\namecite{Cover91} for a discussion of the Shannon-McMillan-Breiman
theorem.}, given a large enough sample, the sample mean of the
negative log probability of a model will converge to its cross entropy
with the true model. That is
\begin{eqnarray}
H(p,q) &=& - \lim_{n \rightarrow \infty} \frac{1}{n} \log
q(w_0^n)
\end{eqnarray}
where $w_0^n$ is a string of the language $L$. In practice, one takes
a large sample of the language, and calculates the negative log 
probability of the sample, normalized by its size\footnote{It is
important to remember to include the end marker in the strings of the
sample.}.  The 
lower the cross entropy (i.e. the higher the probability the model
assigns to the sample), the better the model. 
Usually this is reported in terms of perplexity, which we will do as
well\footnote{When assessing the magnitude of a perplexity
improvement, it is often better to look at the reduction in cross
entropy, by taking the log of the perplexity.  It will be left to the
reader to do so.}.

Some of the trials discussed below will report results in terms of
word and/or sentence error rate, which are obtained when the language
model is embedded in a speech recognition system.  Word error rate is
the number of 
deletion, insertion, or substitution errors per 100 words.  Sentence
error rate is the number of sentences with one or more errors per 100
sentences.

Statistical parsers are typically evaluated for accuracy at the
constituent level, rather than simply whether or not the parse that
the parser found is completely correct or not.  A constituent for
evaluation purposes consists of a label (e.g. NP) and a span
(beginning and ending word positions).  For example, in figure
\ref{fig:itree}(a), there is a VP that spans the words
``\texttt{chased the ball}''.  Evaluation is carried out on a
hand-parsed test corpus, and the manual parses are treated as
correct.   We will call the manual parse GOLD and the parse
that the parser returns TEST.  Precision is
the number of common constituents in GOLD and TEST
divided by the number of constituents in TEST.  Recall is the
number of common constituents in GOLD and TEST
divided by the number of constituents in GOLD. Following
standard practice, we will be reporting scores only for
non-part-of-speech constituents, which are called labeled recall
(LR) and labeled precision (LP).  Sometimes in
figures we will plot their average, and also what can be termed the
parse error, which is one minus their average.  

LR and LP are part of the standard set of {\small PARSEVAL} measures
of parser quality \cite{Black91}.  From this set of measures, we will
also include the crossing bracket scores: average crossing brackets
(CB), percentage of sentences with no crossing brackets (0 CB), and
the percentage of sentences with two crossing brackets or fewer
($\leq$ 2 CB).  In addition, we show the average number
of rule expansions considered per word, i.e. the number of rule
expansions for which a probability was calculated -- see
\namecite{Roark00b} -- and the average number of analyses advanced to
the next priority queue per word. 

This is an incremental parser with a pruning strategy and no
backtracking.  In such a model, it is possible to commit to a set of
partial analyses at a particular point that cannot be completed given
the rest of the input string (i.e. the parser can {\it garden
path\/}).  In such a case, the parser fails to return a complete
parse.  In the event that no complete parse is found, the highest
initially ranked parse on the last non-empty priority queue is returned.  All
unattached words are then attached at the highest level in the tree.
In such a way we predict no new constituents and all incomplete
constituents are closed.  This structure is evaluated for precision
and recall, which is entirely appropriate for these incomplete as well
as complete parses.  If we fail to identify nodes later in the parse,
the recall will suffer, and if our early predictions were bad, both
precision and recall will suffer.  Of course, the percentage of these
failures are reported as well.

\subsection{Parser accuracy and efficiency}
The first set of results looks at the performance of the parser on the
standard corpora for statistical parsing trials: sections 2-21
(989,860 words, 39,832 sentences) of the Penn Treebank \cite{Marcus93}
serving as the training 
data, section 24 (34,199 words, 1,346 sentences) as the held-out
data for parameter estimation, and section 23 (59,100 words, 2,416 
sentences) as the test data. 
Section 22 (41,817 words, 1,700 sentences) served as the development
corpus, on which the parser was tested until stable versions were
ready to run on the test data, to avoid developing the parser to fit
the specific test data.

\begin{table*}[t]
\begin{tabular}{|l|c|c|c|c|c|c|r|c|}
\hline
{\small Conditioning} & {\small LR} & {\small LP} & {\small CB} & 
{\small 0 CB} & {\small $\leq$ 2} & {\small Pct.} &
{\small Avg. rule\ \ } & {\small Average}\\
{} & {} & {} & {} & {} & {\small CB} & {\small failed} & {\small
expansions\ } & {\small analyses}\\
{} & {} & {} & {} & {} & {} & {} & {\small considered${}^{\dag}$} & {\small advanced${}^{\dag}$}\\\hline
\multicolumn{9}{|c|}{section 23: 2245 sentences of length $\leq$ 40}\\\hline
{none} & {71.1} & {75.3} & {2.48} & {37.3} & {62.9} & {0.9} & 
{14,369} & {516.5}\\\hline
{par+sib} & {82.8} & {83.6} & {1.55} & {54.3} & {76.2} & {1.1} & {9,615}
& {324.4}\\\hline
{NT struct} & {84.3} & {84.9} & {1.38} & {56.7} & {79.5} & {1.0} & {8,617}
& {284.9}\\\hline
{NT head} & {85.6} & {85.7} & {1.27} & {59.2} & {81.3} & {0.9} & {7,600}
& {251.6}\\\hline
{POS struct} & {86.1} & {86.2} & {1.23} & {60.9} & {82.0} & {1.0} & {7,327}
& {237.9}\\\hline
{attach} & {86.7} & {86.6} & {1.17} & {61.7} & {83.2} & {1.2} & {6,834}
& {216.8}\\\hline
{all} & {86.6} & {86.5} & {1.19} & {62.0} & {82.7} & {1.3} & {6,379}
& {198.4}\\\hline
\multicolumn{9}{|c|}{section 23: 2416 sentences of length $\leq$ 100}\\\hline
{attach} & {85.8} & {85.8} & {1.40} & {58.9} & {80.3} & {1.5} & {7,210}
& {227.9}\\\hline
{all} & {85.7} & {85.7} & {1.41} & {59.0} & {79.9} & {1.7} & {6,709}
& {207.6}\\\hline
\end{tabular}
\begin{footnotesize}
\ \ \ ${}^{\dag}$per word
\end{footnotesize}
\caption{Results conditioning on various contextual events, standard training and testing corpora}\label{tab:res1}
\end{table*}

Table \ref{tab:res1} shows trials with increasing amounts of
conditioning information from the left-context.  
There are a couple of things to notice from these results.  First, and 
least surprising, is that the accuracy of the parses improved as we
conditioned on more and more information.  Like the non-lexicalized
parser in \namecite{Roark99b}, we found that the search efficiency, in
terms of number of rule expansions considered 
or number of analyses advanced, also improved as we increased the amount
of conditioning.  Unlike the Roark and Johnson parser, however,
our coverage did not substantially drop as the amount of
conditioning information increased, and in some cases improved
slightly.  They did not smooth their conditional probability
estimates, and blamed sparse data for their decrease in coverage as
they increased the conditioning information.  These results appear to
support this, since our smoothed model showed no such tendency.

\begin{figure*}[t]
\begin{center}
\epsfig{file={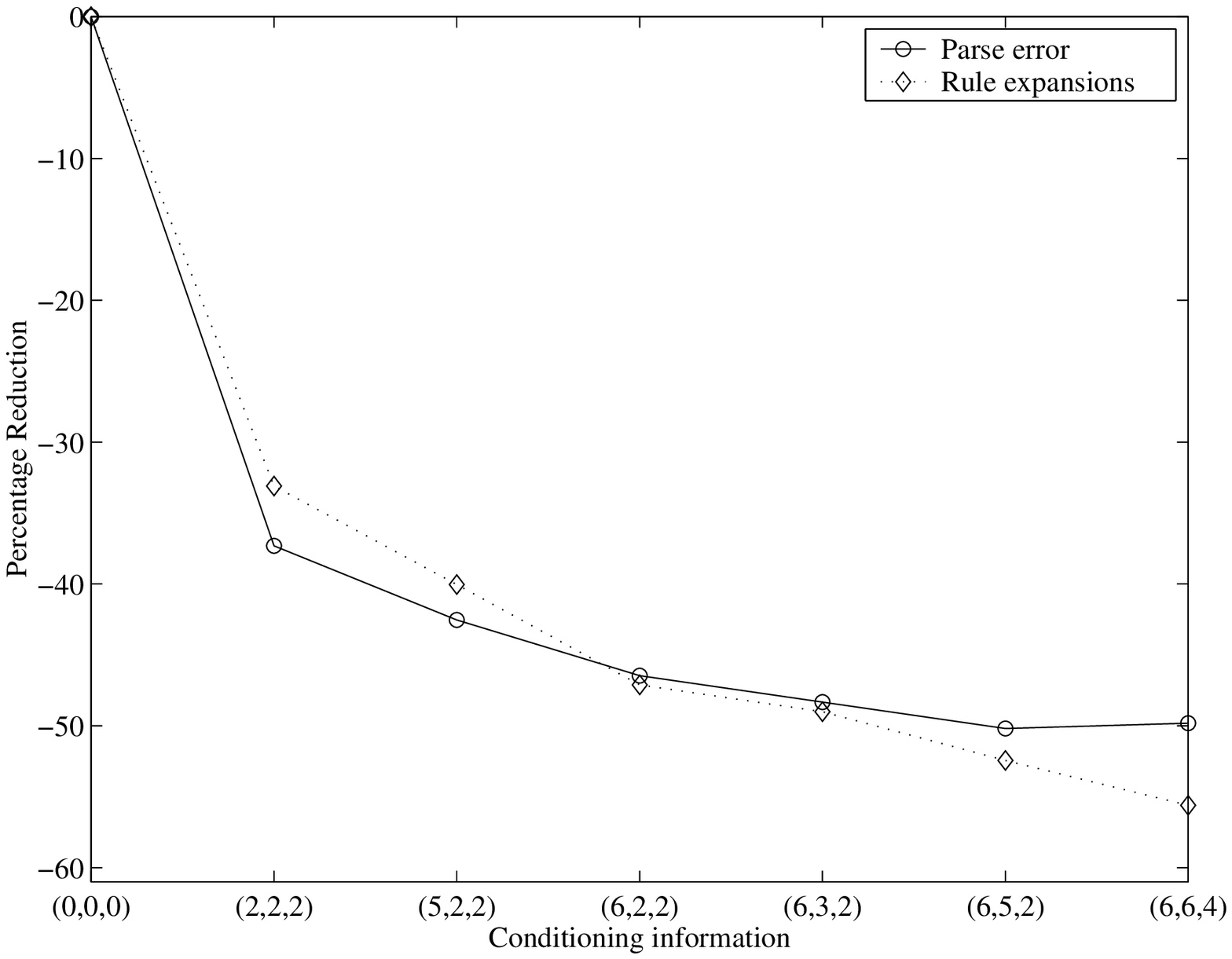}, width = 4in}
\end{center}
\caption{Reduction in average precision/recall error and in number of
rule expansions per word as conditioning increases, for sentences of length
$\leq$ 40} \label{fig:err_red}
\end{figure*}

Figure \ref{fig:err_red} shows the
reduction in parser error, $1-\frac{\mathrm{LR+LP}}{2}$, and 
the reduction in rule expansions considered as the conditioning
information increased.  The bulk of the
improvement comes from simply conditioning on the labels of the parent
and the closest sibling to the node being expanded.  Interestingly,
conditioning all POS expansions on two c-commanding heads
made no accuracy difference compared to conditioning only leftmost
POS expansions on a single c-commanding head; but it did
improve the efficiency.

These results, achieved using very straightforward conditioning events
and considering only the left-context, are within 1-4 points of the
best published accuracies cited above\footnote{Our score of 85.8
average labelled precision and recall for sentences less than or equal
to 100 on section 23 compares to: 86.7 in \namecite{Charniak97}, 86.9
in \namecite{Ratna97}, 88.2 in \namecite{Collins99}, 89.6 in
\namecite{Charniak00}, and 89.75 in \namecite{Collins00}.}.  Of the
2416 sentences in the section, 728 had the totally correct parse, 
30.1 percent tree accuracy.  Also, the parser returns a set of  
candidate parses, from which we have been choosing the top ranked;  if 
we use an oracle to choose the parse with the highest accuracy from
among the candidates (which averaged 70.0 in number per sentence), we
find an average labelled precision/recall of 94.1, for 
sentences of length $\leq$ 100.  The parser, thus,
could be used as a front end to some other model, with the hopes of
selecting a more accurate parse from among the final candidates.

\begin{figure*}[t]
\begin{center}
\epsfig{file=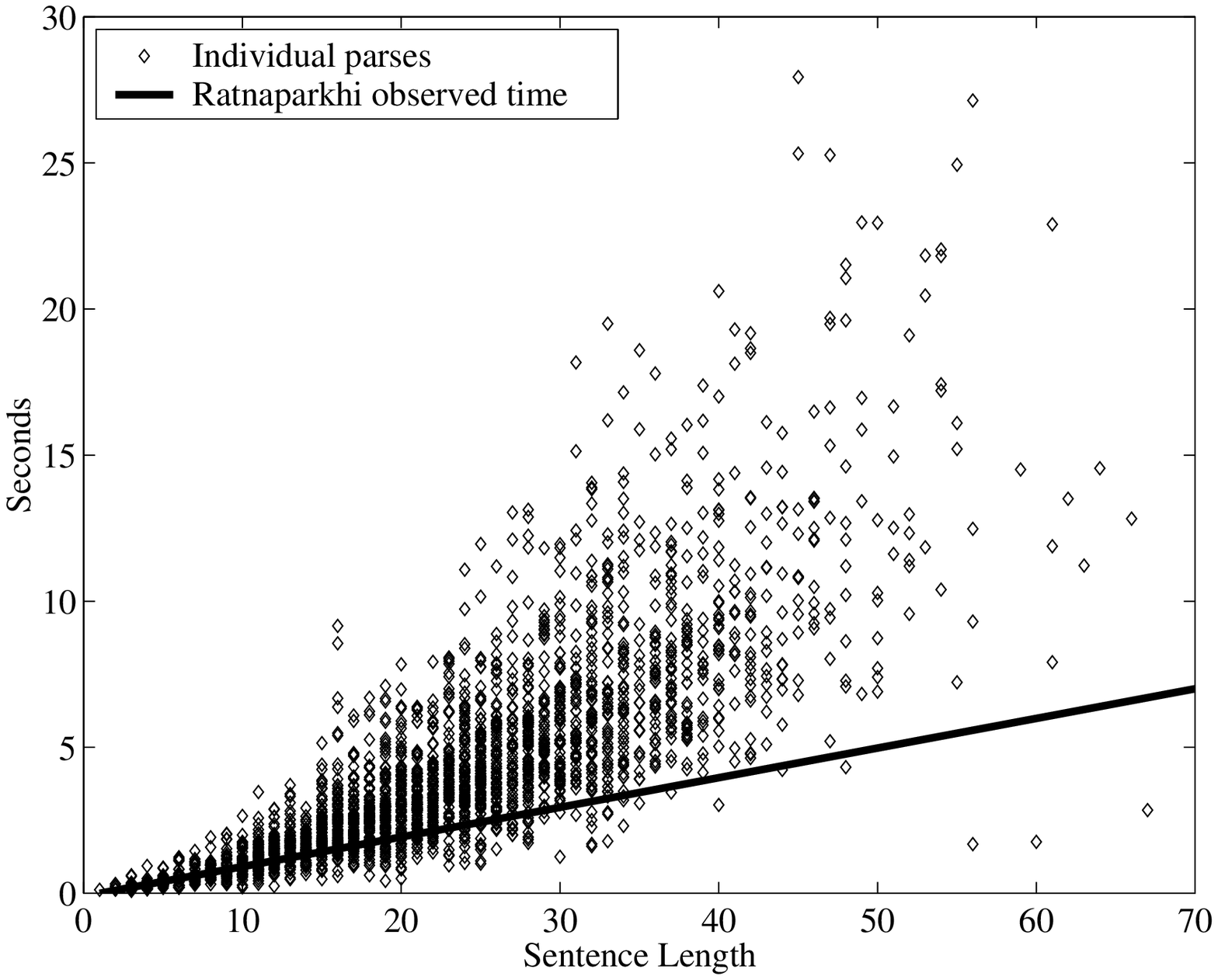, width=4in}
\end{center}
\caption{Observed running time on section 23 of the Penn treebank,
with the full conditional probability model and beam of $10^{-11}$,
using one 300 Mhz UltraSPARC processor and 256MB of RAM of a Sun
Enterprise 450}\label{fig:time}
\end{figure*}

While we have shown that the conditioning information improves the
efficiency in terms of rule expansions considered and analyses
advanced, what does the efficiency of such a parser look like in
practice?  Figure \ref{fig:time} shows the observed time at our
standard base beam of $10^{-11}$ with the full conditioning regimen,
alongside an approximation of the reported observed (linear) time in 
\namecite{Ratna97}.  Our observed times look polynomial, which is to
be expected given our pruning strategy: the denser the competitors
within a narrow probability range of the best analysis, the more time
will be spent working on these competitors; and the farther along in
the sentence, the more chance for ambiguities that can lead to such a
situation.  While our observed times are not linear, and are clearly
slower than his times (even with a faster machine), they are quite
respectably fast.  The differences between a 
k-best and a beam-search parser (not to mention the use of dynamic 
programming) make a running time difference unsurprising. What is
perhaps surprising is that the difference is not greater.
Furthermore, this is quite a large beam (see discussion below),
so that very large improvements in efficiency can be had at the
expense of the number of analyses that are retained.

\pagebreak
\subsection{Perplexity results}
The next set of results will highlight what recommends this approach
most: the ease with which one 
can estimate string probabilities in a single pass from left-to-right
across the string.  By definition, a PCFG's estimate of a
string's probability is the sum of the probabilities of all trees that
produce the string as terminal leaves (see equation \ref{eq:prst}). In 
the beam-search approach outlined above, we can estimate the string's
probability in the same manner, by summing the probabilities of the
parses that the algorithm finds.  Since this is not an exhaustive
search, the parses that are returned will be a subset of the total set 
of trees that would be used in the exact PCFG estimate;
hence the estimate thus arrived at will be bounded above by the
probability that would be generated from an exhaustive search.  The
hope is that a large amount of the probability mass will be accounted
for by the parses in the beam.  The method cannot overestimate the
probability of the string. 

Recall the discussion of the grammar models above, and our
definition of the set of partial derivations $D_{w_0^{j}}$ with
respect to a prefix string $w_0^j$ (see equations \ref{eq:pr_prst}
and \ref{eq:wd_prst}).
By definition, 
\begin{eqnarray}
\Pr(w_{j+1}|w_0^j) &=& \frac{\Pr(w_0^{j+1})}{\Pr(w_0^j)}
\hspace*{.1in}=\hspace*{.1in} \frac{\sum_{d \in
D_{w_0^{j+1}}}\Pr(d)}{\sum_{d \in 
D_{w_0^j}}\Pr(d)}\label{eq:my_prst} 
\end{eqnarray}
Note that the numerator at word $w_j$ is the denominator at word
$w_{j+1}$, so that the product of all of the word probabilities is the 
numerator at the final word, i.e. the string prefix probability.  

We can make a consistent estimate of the string probability by
similarly summing over all of the trees within our beam.  Let
$H_{i}^{init}$ be the priority queue $H_{i}$ before any processing has begun
with word $w_{i}$ in the look-ahead.  This is a subset of the possible 
leftmost partial
derivations with respect to the prefix string $w_0^{i-1}$.
Since $H_{i+1}^{init}$ is produced by expanding only analyses on
priority queue 
$H_{i}^{init}$, the set of complete trees consistent with the partial
derivations on priority queue $H_{i+1}^{init}$ is a subset of the set of
complete trees consistent with the partial derivations on priority queue
$H_{i}^{init}$, i.e. the total probability mass represented by the
priority queues is monotonically decreasing.  Thus conditional word
probabilities defined in a way consistent with equation
\ref{eq:my_prst} will always be between zero and one.  Our conditional 
word probabilities are calculated as follows:
\begin{eqnarray}
\Pr(w_{i}|w_{0}^{i-1}) &=& \frac{\sum_{d\in
H_{i+1}^{init}}\Pr(d)}{\sum_{d\in H_{i}^{init}}\Pr(d)} 
\end{eqnarray}

As mentioned above, the model cannot overestimate the probability 
of a string, because the string probability is simply the sum over the
beam, which is a subset of the possible derivations.  By utilizing a
figure-of-merit to identify promising analyses, we are simply placing
our attention on those parses which are likely to have a high
probability, and thus we are increasing the amount of probability mass
that we do capture, of the total possible.  It is not part of the
probability model itself.  

Since each word
is (almost certainly, because of our pruning strategy) losing some
probability mass, the probability model is not ``proper'', i.e. the
sum of the probabilities over the vocabulary is less than one.  In
order to have a proper probability distribution, we would need to
renormalize by dividing by some factor.  Note, however, that this
renormalization 
factor is necessarily less than one, and thus would uniformly 
increase each word's probability under the model, i.e. any perplexity
results reported below will be higher than the ``true'' perplexity
that would be assigned with a properly normalized distribution.  In
other words, renormalizing would make our perplexity measure lower 
still.  The hope, however, is that the improved parsing model provided
by our conditional probability model will cause the distribution over
structures to be more peaked, thus enabling us to capture more of the
total probability mass, and making this a fairly snug upper bound on
the perplexity.

One final note on assigning probabilities to strings:
because this parser does garden path on a small percentage of
sentences, this must be interpolated with another estimate, to ensure
that every word receives a probability estimate.  In our trials, we
used the unigram, with a very small mixing coefficient:
\begin{eqnarray}
\Pr(w_{i}|w_{0}^{i-1}) &=& \lambda(w_{0}^{i-1})\frac{\sum_{d\in
H_{i+1}^{init}}\Pr(d)}{\sum_{d\in H_{i}^{init}}\Pr(d)} +
(1-\lambda(w_{0}^{i-1}))\Prhat(w_{i}) 
\end{eqnarray}
If $\sum_{d\in H_{i}^{init}}\Pr(d)$ = 0 in our model, then our model
provides no distribution over following words, since the denominator
is zero.  Thus,
\begin{eqnarray}
\lambda(w_{0}^{i-1}) &=& \left\{ \begin{array}{ll} 0 & 
\mathrm{\it if}\ \sum_{d\in H_{i}^{init}}\Pr(d) = 0\\ .999 & otherwise 
\end{array}
\right.
\end{eqnarray}

\begin{table*}[t]
\begin{tabular}{|l|c|c|c|c|c|r|c|}
\hline
{\small Corpora} & {\small Condi-} & {\small LR} & {\small LP} &
{\small Pct.} & {Perplexity} & 
{\small Avg. rule\ \ } & {\small Average}\\
{} & {\small tioning} & {} & {} & {\small failed} & {} & {\small
expansions\ } & {\small analyses}\\
{} & {} & {} & {} & {} & {} & {\small considered${}^{\dag}$} & {\small advanced${}^{\dag}$}\\\hline
\multicolumn{8}{|c|}{sections 23-24: 3761 sentences $\leq$ 120}\\\hline
{unmodified} & {all} & {85.2} & {85.1} & {1.7} & {} & {7,206} &
{213.5}\\\hline
{no punct} & {all} & {82.4} & {82.9} & {0.2} & {} & {9,717} &
{251.8}\\\hline\hline
{C\&J corpus} & {par+sib} & {75.2} & {77.4} & {0.1} & {310.04} & {17,418} &
{457.2}\\\hline
{C\&J corpus} & {{\small NT} struct} & {77.3} & {79.2} & {0.1} & {290.29} & {15,948} &
{408.8}\\\hline
{C\&J corpus} & {{\small NT} head} & {79.2} & {80.4} & {0.1} & {255.85} & {14,239} &
{363.2}\\\hline
{C\&J corpus} & {POS struct} & {80.5} & {81.6} & {0.1} & {240.37} & {13,591} &
{341.3}\\\hline
{C\&J corpus} & {all} & {81.7} & {82.1} & {0.2} & {152.26} & {11,667} &
{279.7}\\\hline
\end{tabular}
\begin{footnotesize}
${}^{\dag}$per word
\end{footnotesize}
\caption{Results conditioning on various contextual events, sections
23-24, modifications following Chelba and Jelinek}\label{tab:res2}
\end{table*}

Chelba and Jelinek \shortcite{Chelba98a,Chelba98b} also used a parser
to help assign word probabilities, via the Structured Language Model
outlined in section \ref{sec:SLM}.  They trained and tested the SLM on
a modified, more ``speech-like'' version of the Penn Treebank.  
Their modifications included: (i) removing orthographic cues to
structure (e.g. punctuation); (ii) replacing all numbers with the
single token {\it N\/}; and (iii) closing the vocabulary at 10,000,
replacing all other words with the UNK token.  They
used sections 00-20 (929,564 words) as the development set, sections
21-22 (73,760 words) as the check set (for interpolation coefficient
estimation), and tested on sections 23-24 (82,430 words).  We obtained
the training and testing corpora from 
them (which we will denote \texttt{C\&J corpus}), and also created 
intermediate corpora, upon which only the first two modifications were
carried out (which we will denote \texttt{no punct}). Differences in
performance will give an indication of the impact on parser 
performance of the different modifications to the corpora.  All trials 
in this section used sections 00-20 for counts, held out 21-22, and
tested on 23-24.

Table \ref{tab:res2} shows several things.  First, it shows relative
performance for unmodified, no punct, and C\&J 
corpora with the full set of conditioning information.  We can see
that removing the punctuation causes (unsurprisingly) a dramatic drop
in the accuracy and efficiency of the parser.  Interestingly, it also
causes coverage to become nearly total, with failure on just two
sentences per thousand on average.  

\begin{figure*}[t]
\begin{center}
\epsfig{file=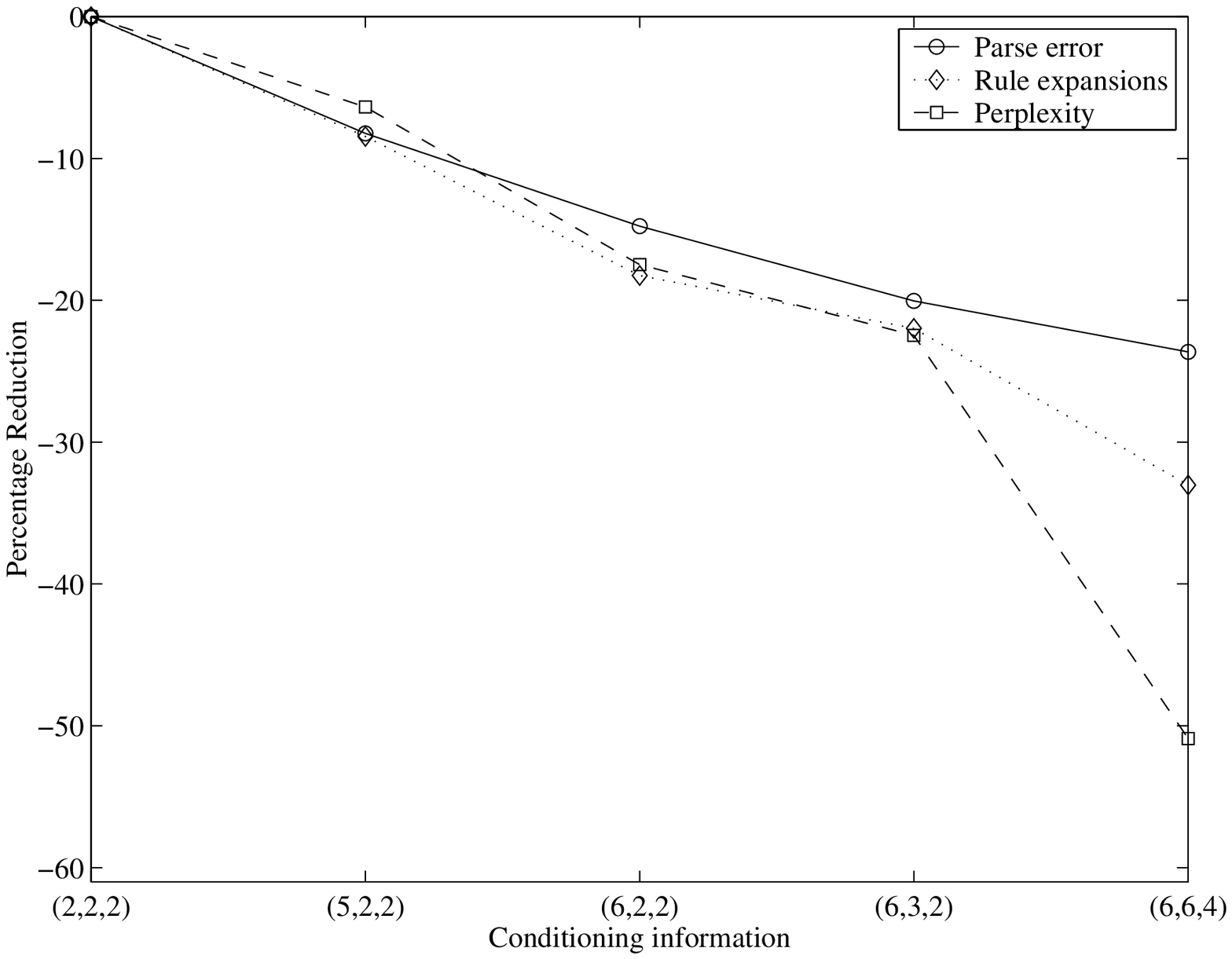, width = 4in}
\end{center}
\caption{Reduction in average precision/recall error, number of
rule expansions, and perplexity as conditioning increases} \label{fig:perp_red}
\end{figure*}

We see the familiar pattern, in the C\&J corpus results, of
improving 
performance as the amount of conditioning information grows.  In this
case we have perplexity results as well, and figure \ref{fig:perp_red} 
shows the reduction in parser error, rule expansions, and perplexity as
the amount of conditioning information grows.  While all three seem to 
be similarly improved by the addition of structural context
(e.g. parents and siblings), the addition of c-commanding heads has
only a moderate effect on the parser accuracy, but a very large effect 
on the perplexity.  The fact that the efficiency was improved more
than the accuracy in this case (as was also seen in figure
\ref{fig:err_red}), seems to indicate that this additional information is
causing the distribution to become more peaked, so that fewer analyses 
are making it into the beam.

Table \ref{tab:res3} compares the perplexity of our model with Chelba
and Jelinek \shortcite{Chelba98a,Chelba98b} on the same training and testing
corpora.  We built an interpolated trigram model to serve as a
baseline (as they did), and also interpolated our model's 
perplexity with the trigram, using the same mixing coefficient as
they did in their trials (taking 36 percent of the estimate from the
trigram)\footnote{Our optimal mixture level was closer to 40 percent,
but the difference was negligible.}.  The trigram model was also
trained on sections 00-20 of the C\&J corpus.  Trigrams and bigrams
were binned by the total count of the conditioning words in the
training corpus, and maximum likelihood mixing coefficients were
calculated for each bin, to mix the trigram with bigram and unigram
estimates.  Our trigram model performs at almost exactly the 
same level that theirs does, which is what we would expect.  Our
parsing model's perplexity improves upon their first result 
fairly substantially\footnote{Recall, that our perplexity measure
should, ideally, be even lower still.}, but is only slightly better
than their second 
result.  However, when we interpolate with the trigram, we see that
the additional improvement is greater than the one they experienced.
This is not surprising, since our conditioning information is in many
ways orthogonal to that of the trigram, insofar as it includes the
probability mass of the derivations;  in contrast, their model in
some instances is very close to the trigram, by conditioning on two
words in the prefix string, which may happen to be the two adjacent
words.

\begin{table*}[t]
\begin{tabular} {|p{1.9in}|p{.9in}|p{.9in}|p{.9in}|}
\hline
{\small Paper} & \multicolumn{3}{|c|}{Perplexity}\\\cline{2-4}
{} & {\small Trigram Baseline} &
{\small Model} & {\small Interpolation, $\lambda$=.36}\\\hline
{\namecite{Chelba98a}} & {167.14} & {158.28} & {148.90}\\\hline
{\namecite{Chelba98b}} & {167.14} & {153.76} & {147.70}\\\hline
{Current results} & {167.02} & {152.26} & {137.26}\\\hline
\end{tabular}
\caption{Comparison with previous perplexity results}\label{tab:res3}
\end{table*}

These results are particularly remarkable, given that we did not
build our model as a language model {\it per se\/}, but rather as a
parsing model.  The perplexity improvement was achieved by simply taking the
existing parsing model and applying it, with no extra training beyond
that done for parsing.

The hope was expressed above that our reported perplexity would be
fairly close to the ``true'' perplexity that we would achieve if the
model were properly normalized, i.e. that the amount of probability
mass that we lose by pruning is small.  One way to test this is the
following\footnote{Thanks to Ciprian Chelba for this suggestion.}: at
each point in the sentence, calculate the conditional probability of
each word in the vocabulary given the previous words, and sum them.
If there is little loss of probability mass, the sum should be close
to one.  We did this for the first 10 sentences in the test corpus, a
total of 213 words (including the end-of-sentence markers).  One of
the sentences was a failure, so that 12 of the word probabilities (all
of the words after the point of the failure) were not estimated by our
model.  Of the remaining 201 words, the
average sum of the probabilities over the 10,000 word vocabulary was
0.9821, with a minimum of 0.7960, and a maximum of 0.9997.
Interestingly, at the word where the failure occurred, the sum of the
probabilities was 0.9301.

\subsection{Word error rate}
In order to get a sense of whether these perplexity reduction results
can translate to improvement in a speech recognition task, we
performed a very small preliminary experiment on N-best lists.  The
DARPA `93 HUB1 test 
setup consists of 213 utterances read from the Wall St. Journal, a
total of 3446 words.  The corpus comes with a baseline trigram model,
using a 20,000 word open vocabulary, and trained on approximately 40
million words.  We used Ciprian Chelba's A$^\star$
decoder\footnote{See \namecite{Chelba00} for details.} to find the
50 best hypotheses from each lattice, along with the acoustic and
trigram scores.  Given the idealized circumstances of the
production (text read in a lab), the lattices are relatively sparse,
and in many cases 50 distinct string hypotheses were not found in a
lattice.  We reranked an average of 22.9 hypotheses with our
language model per utterance.

\begin{table*}[t]
\begin{tabular} {|p{1.3in}|p{.6in}|p{.7in}|p{.4in}|p{.7in}|p{.6in}|}
\hline
{Model} & {Training Size} & {Vocabulary Size} & {LM Weight} &
{Word Error Rate \%} & {Sentence Error}\\
{} & {} & {} & {} & {} & {Rate \%} \\\hline 
{Lattice trigram} & {40M} & {20K} & {16} & {13.7} & {69.0}\\\hline
{\namecite{Chelba00} ($\lambda$=.4)} & {20M} & {20K} & {16} & {13.0} &
{}\\\hline 
{Current model} & {1M} & {10K} & {15} & {15.1} & {73.2}\\\hline
{Treebank trigram} & {1M} & {10K} & {5} & {16.5} & {79.8}\\\hline
{No language model} & {} & {} & {0} & {16.8} & {84.0}\\\hline
\end{tabular}
\caption{Word and sentence error rate results for various models, with
differing training and vocabulary sizes, for the best language model
factor for that particular model} \label{tab:wer}
\end{table*}

One complicating issue has to do with the tokenization in the Penn
Treebank versus that in the HUB1 lattices.  In particular,
contractions (e.g. \texttt{he's}) are split in the Penn Treebank
(\texttt{he 's}) but not in the HUB1 lattices.  Splitting of the
contractions is critical for parsing, since the two parts oftentimes
(as in the previous example) fall in different constituents.  We
follow \namecite{Chelba00} in dealing with this problem: for parsing
purposes, we use the Penn Treebank tokenization;  for interpolation
with the provided trigram model, and for evaluation, the lattice
tokenization is used.  If we are to interpolate our model with the
lattice trigram, we must wait until we have our model's estimate for
the probability of both parts of the contraction; their product can
then be interpolated with the trigram estimate.  In fact, 
interpolation in these trials made no improvement over the better of
the uninterpolated models, but simply resulted in performance
somewhere between the better and the worse of the two models, so we
will not present interpolated trials here.

Table \ref{tab:wer} reports the word and sentence error rates for
five different models:  (i) the trigram model that comes with the
lattices, trained on approximately 40M words, with a vocabulary of
20,000; (ii) the best performing model from \namecite{Chelba00}, which
was interpolated with the lattice trigram at $\lambda$=0.4; (iii) 
our parsing model, with the same training and vocabulary as the
perplexity trials above; (iv) a trigram model with the same training
and vocabulary as the parsing model; and (v) no language model at all.
This last model shows the performance from the acoustic model
alone, without the influence of the language model.  The log of the
language model score is multiplied by the language model (LM)
weight when summing the logs of the language and acoustic scores, as a
way of increasing the relative contribution of the language model to
the composite score.  We followed \namecite{Chelba00} in using an
LM weight of 16 for the lattice trigram.  For our model and the
treebank trigram model, the LM weight that resulted in the
lowest error rates is given.

The small size of our training 
data, as well as the fact that we are rescoring N-best lists, rather
than working directly on lattices, makes comparison with the other
models not particularly informative.  What is more informative is the
difference between our model and the trigram trained on the same
amount of data.  We achieved an 8.5 percent relative improvement in
word error rate, and an 8.3 percent relative improvement in sentence
error rate over the treebank trigram.  Interestingly, as mentioned
above, interpolating two models 
together gave no improvement over the better of the two, whether our
model was interpolated with the lattice or the treebank trigram.  This
contrasts with our perplexity results reported above, as well as with
the recognition experiments in \namecite{Chelba00}, where the best
results resulted from interpolated models.  

The point of this small experiment was to see if our parsing model could
provide useful information even in the case that recognition errors
occur, as opposed to the (generally) fully grammatical strings upon which
the perplexity results were obtained.  As one reviewer pointed out, given
that our model relies so heavily on context, it may have difficulty
recovering from even one recognition error, perhaps more difficulty
than a more locally-oriented trigram.  While the improvements over the
trigram model in 
these trials are modest, they do indicate that our model is robust
enough to provide good information even in the face of noisy input.
Future work will include more substantial word recognition experiments. 

\subsection{Beam variation}
The last set of results that we will present addresses the question of
how wide the beam must be for adequate results.  The base beam factor
that we have used to this point is $10^{-11}$, which is quite wide.
It was selected with the goal of high parser accuracy; but in this new 
domain, parser accuracy is a secondary measure of performance.  
To determine the effect on perplexity, we
varied the base beam factor in trials on the Chelba and Jelinek
corpora, keeping the level of conditioning
information constant, and table \ref{tab:res4} shows the
results across a variety of factors.

\begin{table*}[t]
\begin{tabular}{|l|c|c|c|c|c|r|r|}
\hline
{\small Base} & {\small LR} & {\small LP} &
{\small Pct.} & {Perplexity} & {Perplexity} & 
{\small Avg. rule\ \ } & {\small Words per}\\
{\small Beam} & {} & {} & {\small failed} & {\small $\lambda$=0} &
{\small $\lambda$=.36} & {\small expansions\ } & {\small second\ \ \ \ }\\
{\small Factor} & {} & {} & {} & {} & {} & {\small
considered${}^{\dag}$} & {}\\\hline 
\multicolumn{8}{|c|}{sections 23-24: 3761 sentences $\leq$ 120}\\\hline
{\small $10^{-11}$} & {81.7} & {82.1} & {0.2} & {152.26} & {137.26} & {11,667} &
{3.1}\\\hline
{\small $10^{-10}$} & {81.5} & {81.9} & {0.3} & {154.25} & {137.88} & {6,982} &
{5.2}\\\hline
{\small $10^{-9}$} & {80.9} & {81.3} & {0.4} & {156.83} & {138.69} & {4,154} &
{8.9}\\\hline
{\small $10^{-8}$} & {80.2} & {80.6} & {0.6} & {160.63} & {139.80} & {2,372} &
{15.3}\\\hline
{\small $10^{-7}$} & {78.8} & {79.2} & {1.2} & {166.91} & {141.30} & {1,468} &
{25.5}\\\hline
{\small $10^{-6}$} & {77.4} & {77.9} & {1.5} & {174.44} & {143.05} & {871} &
{43.8}\\\hline
{\small $10^{-5}$} & {75.8} & {76.3} & {2.6} & {187.11} & {145.76} & {517} &
{71.6}\\\hline
{\small $10^{-4}$} & {72.9} & {73.9} & {4.5} & {210.28} & {148.41} & {306} &
{115.5}\\\hline
{\small $10^{-3}$} & {68.4} & {70.6} & {8.0} & {253.77} & {152.33} & {182} &
{179.6}\\\hline
\end{tabular}
\begin{footnotesize}
\ \ \ \ \ ${}^{\dag}$per word
\end{footnotesize}
\caption{Results with full conditioning on the C\&J corpus
at various base beam factors}\label{tab:res4}
\end{table*}

\begin{figure*}[t]
\begin{center}
\epsfig{file=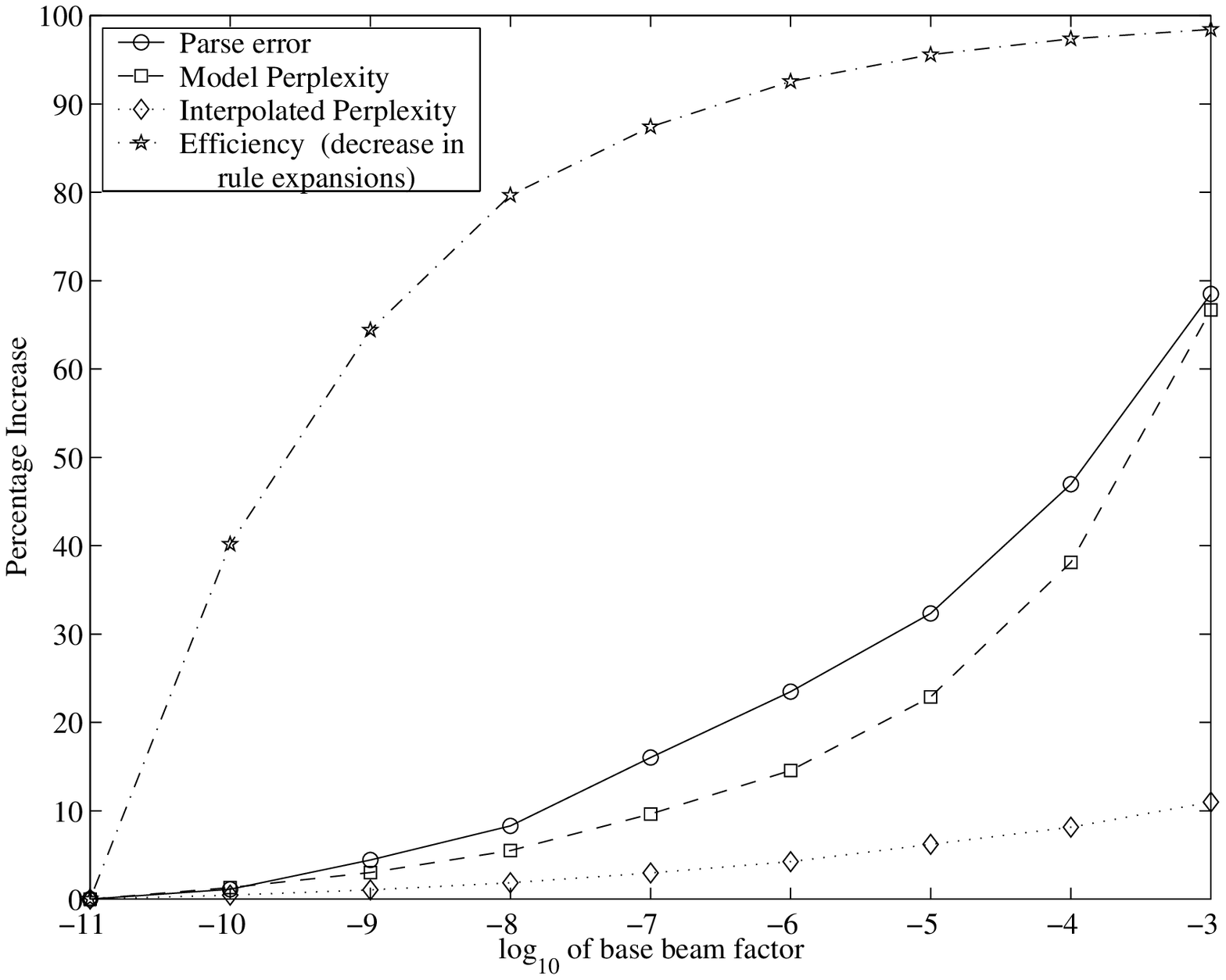, width = 4in}
\end{center}
\caption{Increase in average precision/recall error, model perplexity, 
interpolated perplexity, and efficiency (i.e. decrease in rule expansions
per word) as base beam factor decreases} \label{fig:beam}
\end{figure*}

The parser error, parser coverage, and the uninterpolated model
perplexity ($\lambda$ = 1) all suffered substantially from a narrower
search,  
but the interpolated perplexity remained quite good even at the
extremes.  Figure \ref{fig:beam} plots
the percentage increase in parser error, model perplexity,
interpolated perplexity, and efficiency (i.e. decrease in rule expansions
per word) as the base beam factor decreased.  Note that the model
perplexity and parser accuracy are quite similarly affected, but that
the interpolated perplexity remained far below the trigram baseline,
even with extremely narrow beams.

\section{Conclusion and Future Directions}
The empirical results presented above are quite encouraging, and the
potential of this kind of approach both for parsing and language
modeling seems very promising.  With a simple 
conditional probability model, and simple statistical search
heuristics, we were able to find very accurate parses efficiently,
and, as a side effect, were able to assign word probabilities that
yield a perplexity improvement over previous results.  These
perplexity improvements are particularly promising, because the parser 
is providing information that is, in some sense, orthogonal to the
information provided by a trigram model, as evidenced by the robust
improvements to the baseline trigram when the two models are
interpolated.  

There are several important future directions that will be taken in
this area.  First, there is reason to believe that some of the conditioning
information is not uniformly useful, and we would benefit from finer
distinctions.  For example, the probability of a preposition is presumably
more dependent on a c-commanding head than the probability of a
determiner is.  Yet in the 
current model they are both conditioned on that head, as leftmost
constituents of their respective phrases.  Second, there are
advantages to top-down parsing that have not been examined
to date, e.g. empty categories.  A top-down parser, in contrast to a
standard bottom-up chart parser, has enough information to predict
empty categories only where they are likely to occur.  By including
these nodes (which are in the original annotation of the Penn
Treebank), we may be able to bring certain long distance dependencies into a
local focus.  In addition, as mentioned above, we would like to
further test our language model in speech recognition tasks, to see if
the perplexity improvement that we have seen can lead to
significant reductions in word error rate.

Other parsing approaches might also be used in the way that we have
used a top-down parser.
Earley and left-corner parsers, as mentioned in the introduction, also
have rooted derivations that can be used to calculated generative
string prefix probabilities incrementally.  In fact, left-corner
parsing can be simulated by a top-down parser by transforming the
grammar, as was done in \namecite{Roark99b}, and so an approach very
similar to the one outlined here 
could be used in that case.  Perhaps some compromise between the fully
connected structures and extreme underspecification will yield an
efficiency improvement.  Also, the advantages of head-driven parsers
may outweigh their inability to interpolate with a trigram, and lead
to better off-line language models than those that we have presented
here. 

Does a parsing model capture exactly what we need for informed
language modeling?  The answer to that is no.  Some information is
simply not structural in nature (e.g. topic), and we might expect
other kinds of models to be able to better handle non-structural
dependencies.  The improvement that we derived from interpolating the
different models above indicates that using multiple models may be the
most fruitful path in the future.  In any case, a parsing model of the
sort that we have presented here should be viewed as an important potential
source of key information for speech recognition.  Future research
will show if this early promise can be fully realized.\vspace*{.2in}

\starttwocolumn
\begin{acknowledgments}
The author wishes to thank Mark Johnson for invaluable discussion,
guidance and moral support over the course of this project.  Many
thanks also to Eugene Charniak for the use of certain grammar training
routines, and for an enthusiastic interest in the project.  Thanks
also to four anonymous reviewers for valuable and insightful comments,
and to Ciprian Chelba, Sanjeev Khudanpur, and Frederick Jelinek for
comments and suggestions.  Finally, the author would like to
express his appreciation to the participants of discussions during
meetings of the Brown Laboratory for Linguistic Information Processing
(BLLIP); in addition to Mark and Eugene:  Yasemin Altun, Don Blaheta,
Sharon Caraballo, Massimiliano Ciaramita, Heidi Fox, Niyu Ge, and Keith
Hall.  This research was supported in part by NSF IGERT Grant
\#DGE-9870676.
\end{acknowledgments} 

\bibliographystyle{fullname}
\bibliography{cl-00-17}
\end{document}